\documentclass[a4paper]{article}

\newcommand{\comment}[1]{}
\usepackage[square]{natbib}
\let\cite\citep
\usepackage{graphicx}
\usepackage{xcolor}
\usepackage{rotating}
\usepackage{pdflscape}

\usepackage{hyperref}
\hypersetup{
colorlinks = true,
linkcolor = {green!30!black},
citecolor = {green!30!black},
}

\makeatletter

\DeclareRobustCommand\citepos
{\begingroup
\let\NAT@nmfmt\NAT@posfmt
\NAT@swafalse\let\NAT@ctype\z@\NAT@partrue
\@ifstar{\NAT@fulltrue\NAT@citetp}{\NAT@fullfalse\NAT@citetp}}

\let\NAT@orig@nmfmt\NAT@nmfmt
\def\NAT@posfmt#1{\NAT@orig@nmfmt{#1's}}

\makeatother

\usepackage{amssymb,amsmath,mathtools}
\DeclareMathOperator*{\argmin}{\arg\!\min}

\newcommand{\dpar}[2]{\frac{\partial #1}{\partial #2}}

\newcommand{\params}{\ensuremath{\boldsymbol\theta}}
\newcommand{\oldparams}{\ensuremath{\params_{\mathit{old}}}}
\newcommand{\fatx}{\ensuremath{\mathbf{x}}}
\newcommand{\faty}{\ensuremath{\mathbf{y}}}

\newcommand{\Dcal}{\ensuremath{\mathcal{D}}}
\newcommand{\Tcal}{\ensuremath{\mathcal{T}}}
\newcommand{\Xcal}{\ensuremath{\mathcal{X}}}
\newcommand{\Ycal}{\ensuremath{\mathcal{Y}}}
\newcommand{\Zcal}{\ensuremath{\mathcal{Z}}}

\DeclarePairedDelimiter\norm{\lVert}{\rVert}%

\usepackage{tikz}
\usetikzlibrary{decorations.markings}

\usepackage{booktabs}

\author{Maarten Grachten\\Carlos Eduardo Cancino Chac\'on}
\title{Strategies for Conceptual Change in Convolutional Neural Networks}

\date{}

\begin{document}
\maketitle

\begin{center}
Austrian Research Institute for Artificial Intelligence
\end{center}

\begin{center}
\small{OFAI-TR-2015-04 Version 1.1}
\end{center}

\begin{abstract}
A remarkable feature of human beings is their capacity for creative behaviour, referring to their ability to react to problems in ways that are novel, surprising, and useful.
Transformational creativity is a form of creativity where the creative behaviour is induced by a transformation of the actor's \emph{conceptual space}, that is, the representational system with which the actor interprets its environment.
In this report, we focus on ways of adapting systems of learned representations as they switch from performing one task to performing another.
We describe an experimental comparison of multiple strategies for adaptation of learned features, and evaluate how effectively each of these strategies realizes the adaptation, in terms of the amount of training, and in terms of their ability to cope with restricted availability of training data.
We show, among other things, that across handwritten digits, natural images, and classical music, adaptive strategies are systematically more effective than a baseline method that starts learning from scratch.
\end{abstract}

\newpage

\tableofcontents

\newpage

\section{Introduction}
\label{sec:introduction}
In order to survive, any living organism has to adapt to its environment.
Far from being static, the environment for most organisms is subject to constant change.
Arguably, complex and plastic information processing structures such as the mammal brain are an evolutionary answer to the requirement to adapt to unforeseen circumstances during the lifespan of the organism~\cite{allen2012lives}.

Some aspects of this adaptive behaviour, most notable in higher mammals such as humans, are called \emph{creative}.
Although the term is notoriously evasive of a precise definition, there is common agreement that creative behaviour involves elements such as novelty, surprise, and value \cite{weisberg1993creativity,csikszentmihalyi1996creativity}.
An example of creative behaviour can be found in the use of domain names in the world wide web.
Although Top Level Domains (such as \emph{.com}, \emph{.org}, \emph{it}) are intended as a means of structuring the world wide web, they are nowadays used for their linguistic meaning, rather than their original denotation of thematical or geographical structure.
Examples of such uses are \emph{youtu.be}, \emph{friend.ly}, and \emph{podca.st}.
Arguably, this creative use is driven by the increasing use and communication of URLs, and therefore a growing need for shorter URLs, that are easy to memorize.
%


One of the most seminal attempts at formalizing the notion of creativity is that of \citet{Boden:2004wl}.
She links creative behaviour of an agent to its \emph{conceptual space}.
This space represents the way the agent interprets its perception and actions.
According to Boden, creativity amounts to mapping, exploration, and transformation of a conceptual space by an agent.
An example of such a transformation given by \citet[p.
522]{boden1994precis} is the transformation of the conceptual space for music by Arnold Schoenberg, who invented radically new music by dropping the constraint of the home key from the existing conceptual space of music.

Most literature on creativity focuses on the high level, cognitive phenomena it involves, taking for granted that a conceptual space, a way of dealing with the environment, is already present.
In contrast to viewing creativity as a discrete transformation of one static conceptual space into another, a promising alternative perspective is to regard conceptual spaces as inherently dynamic.
In this view, creativity is an inherent property of the dynamics of conceptual spaces that brings them into existence in the first place.
This view also suggests a more gradual distinction between perception and cognition.
\citet[p.
308]{hofstadter2008fluid} goes so far as to say that the ability to \emph{reperceive} is a critical element of creativity.

This focus on the dynamic aspects of conceptual spaces naturally leads to \emph{representation learning}~\cite{bengio13:_repres_learn}, an active field of research in machine learning.
In this area, computational models, predominantly in the form of neural networks, are trained on data to form hierarchically structured representations of the data.
Often, such representations capture semantically relevant characteristics of the data at several levels, and as such form a robust basis for subsequent tasks such as classification, or organization of data.

Typical methods for representation learning deal with the classical machine learning scenario in which a model is trained on a set of data $\fatx$, possibly with labels $\faty$, drawn from a distribution $p(\fatx, \faty)$ (the generating distribution), and evaluated on further data that is also assumed to be from the same distribution $p(\fatx, \faty)$.
This implies that the training methods are designed to converge to a single set (or hierarchical structure) of representations, that is optimal given $p(\fatx, \faty)$.
As stated at the beginning of this Section however, creativity is often driven by a need to adapt to a changing environment.
In terms of the machine learning paradigm, this violates the assumption that $p(\fatx, \faty)$ is static over time.

Thus, for a representation learning model to stay effective in the face of a changing environment requires methods beyond standard representation learning algorithms.
This shift of focus from learning representations in a static environment to adapting learned representations in a dynamic environment brings on multiple non-trivial problems to be addressed.
For example, there is a need for a measure of how well a given representation suits the environmental requirements.
Furthermore, in the light of an environmental change, sometimes gradual change of the learned representations may be beneficial, whereas on other occasions, a more radical change (for example by learning representations from scratch) may be more effective.

It is not in the scope of the current report to address all of these problems.
In this document, we restrict ourselves to the scenario where a radical change in environment is given, and investigate which of several alternative strategies for dealing with that change are most effective in neural network models for representation learning.
We are ultimately interested in finding generally useful strategies that allow a computational system to adapt efficiently to changing environmental requirements.
We start by a precise description of the problem, and the model architecture we use for learning.
Then we define a number of possible adaptation strategies, and report on an experiment in which these strategies are compared by evaluating them on several data sets.
Before that, we discuss how the work presented here relates to creativity in a broader sense.

\subsection{From adaptive representations to creative behaviour}\label{sec:from-adapt-repr}
In this document, we focus on strategies for conceptual change to adapt learned representations from one task to another task.
The tasks we consider here are classification, and autoencoding.
A change of task (for the model to adapt to) may either mean switching from classification to auto-encoding or vice versa, or in the case of classification, a change of target classes.
Note however, that the change of task inducing the conceptual change (the changes in the learned representations) in these cases is an extrinsic factor.
From the perspective of an agent interacting with an outside world, such changes may reflect changes in its environment (e.g.
a change of habitat, with different species and objects).
This scenario corresponds to the type of creativity described above, a form of adaptive, problem solving behaviour.
A different, but related interpretation emphasizes the generative, aesthetic aspects of creative behaviour, that may lead to novel artefacts.
In this case, the conceptual change manifest in the creative behaviour is thought to be driven by intrinsic factors rather than as a response to changes in the environment.

Are the methods for conceptual change described here, and evaluated in the context of classification and autoencoding, also of use in the realization of this second kind of creativity?
Two current computational theories of creative behaviour suggest they may be, since both posit that at the basis of creative behaviour are mechanisms that encode incoming information, and adapt to a dynamic environment to provide optimal encodings, in line with evidence for learning principles in living organisms, such as \emph{minimum entropy coding} \cite{barlow89:_unsup_learn,atick11:_could,olsh96}.
We will briefly describe both theories.

\citet{wiggins15:_idyot} propose a theory of creativity based on the Global Workspace Theory~\cite{baars2002conscious}.
The criterion that determines what representations develop in this model is based on information-theoretic principles, demanding that the representations facilitate accurate prediction of the perceptual future, leading to more efficient information processing.
In their model, multiple competing \emph{generators} match perceptual input to learned representations from memory to predict future input. The representations most of the effective generators surface into the \emph{Global Workspace}, reflecting \emph{conscious awareness}.
This awareness in turn, updates an associative memory that stores the adaptive representations to be used by the generators.
The changes in semantics as a result of adapting the conceptual space (for instance in the form of \emph{aberration} \cite{wiggins06:_prelim_framew_descr_analy_compar_creat_system}) may be regarded as a restricted form of creative behaviour. In addition, \citet{wiggins15:_idyot} theorize that an instantiation of the theory in the form of a computational model, after being exposed to stimuli that have imprinted memories, is able to generate novel artefacts: In the absence of external stimuli, the system can fill its perceptual buffers from memory, thus continuing the cycle of generation from memory, selection, and updating memory.

\citet{schmidhuber10:_formal_theor_creat_fun_intrin_motiv} also argues that a fundamental aspect of cognition in an agent is to learn to compress (or equivalently, predict), the outside world and its affordances.
Plausible drivers for this goal of optimal compression/prediction are extrinsic rewards such as sparing use of physiological resources (information that can be sparsely represented takes less energy to process and retain), and a selective advantage (by allowing better anticipation of future events).
But Schmidhuber argues that an important explanatory factor for human behaviour is an \emph{intrinsic} motivation to \emph{improve} compression/prediction capabilities.
This means that for an agent it is desirable in itself to discover ways of better compressing and predicting its experience, where the (intrinsic) reward is proportional to the degree of improvement.
Schmidhuber claims that this intrinsic motivation manifests as \emph{curiosity} in human beings, and that the subjective experience of improving compression/prediction is aesthetic (that is: fun, beauty, or elegance).
This also leads to a notion of surprise that is different from the usual information-theoretic notion of surprise, in the sense that rather than the information content of an event itself, it is the degree to which the information content of an event triggers an improvement of the agent's compression/predicting capabilities.
Thus, just as a sequence of repeated events (no information content), a sequence of random events (high information-content) is not surprising, and will soon become boring.

Both of the theories of creative behaviour described above involve an adaptive component that generates, predicts, or compresses perceptual input, in other words, it maps incoming data to an internal representation that adapts to accommodate novel patterns in the data.
The precise nature of this accommodation process is not part of the above (rather high level) theories, but the methods evaluated here hint at possible implementations of such a process, especially in situations where the patterns in the data change drastically.
Furthermore, the fact that the methods proposed here concern (deep) neural networks suits the \emph{reinforcement learning} (RL) paradigm of Schmidhubers theory, where neural networks in combination with Q-learning \cite{watkins1989learning} have been shown to be very successful at a variety of game-playing tasks~\cite{2013arXiv1312.5602M,2015arXiv150906461V}.




The remainder of this document is structured as follows.
In Section~\ref{sec:related-work}, we discuss a number of related problems known from machine learning and discuss their relevance for the problem at hand.
In Section~\ref{sec:method}, we introduce the architecture of the neural network model we use, and a number of possible adaptation strategies to facilitate conceptual change.
Section~\ref{sec:experiment} presents the setup of the comparative evaluation of the strategies.
The results of this evaluation are reported, and discussed in Section~\ref{sec:resanddisc}.
Conclusions are presented in Section~\ref{sec:conclusions}.


\section{Related work}\label{sec:related-work}


\subsection{Transfer learning}\label{sec:transfer-learning}
Transfer learning is a sub-field of machine learning that deals with the question how to wield knowledge obtained for some task in some domain (the \emph{source} task and domain, respectively), to better solve another task in a (possibly different) domain (the \emph{target} task and domain, respectively).
This rather general definition leaves room for a number of variations of transfer learning problems, depending on the conditions~\cite{PanY09TKDE}. Most of these transfer learning problems can be characterized with the help of a probabilistic formalization: 
Let \Xcal{} be a feature space, and $P(X)$ ($X \in \Xcal$) a marginal distribution over \Xcal{}. Then the tuple $\Dcal = \{\Xcal, P(X)\}$ is called a \emph{domain}. Furthermore, let \Ycal{} be a label space and $P(Y|X)$ ($X \in \Xcal, Y \in \Ycal$) a conditional distribution over \Ycal{}, given X. Then the tuple $\Tcal = \{\Ycal, P(Y|X)\}$ is called a \emph{task}\footnote{The task specifies the marginal distribution $P(Y|X)$ to be approximated}.
Where necessary, the sub-scripts $s$ and $t$ are used on any of these elements to denote that the elements belong to the \emph{source} and \emph{target} domains, respectively.

\emph{Domain adaptation} (see section~\ref{sec:domain-adaptation}) is a form of transfer learning that assumes the same task, but a different domain (i.e. $\Dcal_s \neq \Dcal_t$)), whereas in \emph{inductive transfer} the essential element is that the task is different (i.e. $\Tcal_s \neq \Tcal_t$)~\cite{ben-david07:_analy, PanY09TKDE}.
Another sub-problem of transfer learning is \emph{class imbalance}, where a particular class is substantially under-represented in the source domain, with respect to the target domain, or vice versa: $P_s(X \mid Y) \neq P_t(X \mid Y)$ \cite{jiang2008literature}.
Finally, \emph{covariate shift} refers to the situation where the relation of a class label to the feature space changes from the source to the target domain: $P_s(Y \mid X) \neq P_t(Y \mid X)$ \cite{jiang2008literature}.
This problem is also known as \emph{concept drift} (see Section~\ref{sec:concept-drift}), with the difference that the latter term is more commonly used in contexts of online learning~\cite{widmer96:_learn}: rather than dealing with distinct source and target domains/tasks, concept drift refers to a gradual change of $P(Y \mid X)$ over time within a domain.

\begin{figure}[t]
\centering
\includegraphics[width=.5\linewidth]{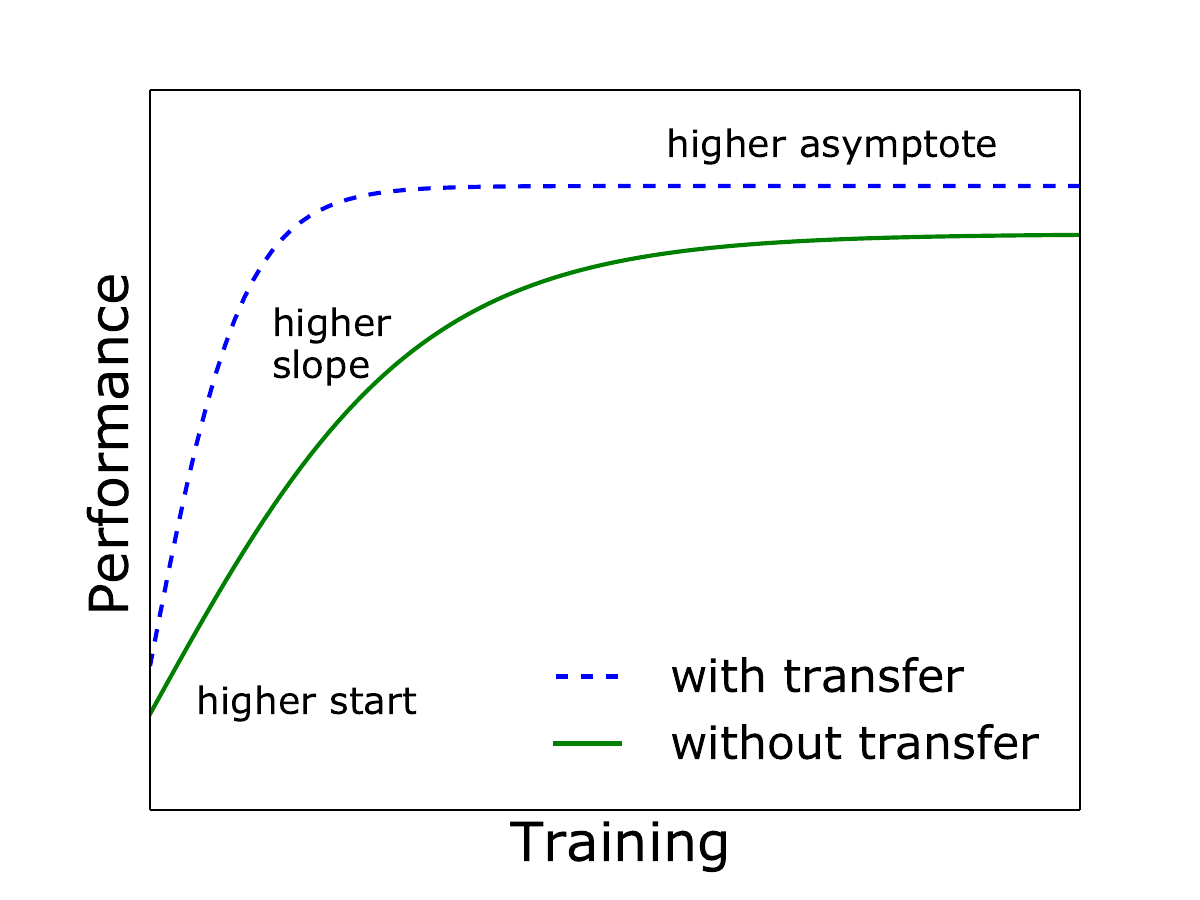}
\caption{Three possible ways in which transfer learning may enhance performance on the target task. Adapted from \protect{\cite{torrey09:_trans}}
}
\label{fig:transfer_plot}
\end{figure}

\citet{torrey09:_trans} describe three ways in which transfer from a source task/domain can help a model perform a target task, as illustrated in Figure~\ref{fig:transfer_plot}.
Most studies concerning transfer loss use only the asymptotic performance as a criterion for successful transfer.
A common measure is the \emph{transfer-loss} \cite{ICML2011Glorot_342}.
It measures the loss of the transfer method minus the loss of an in-domain baseline method for the target domain, that is not informed by the source domain.
The transfer-loss is thus proportional to the difference in asymptotic performance in Figure~\ref{fig:transfer_plot}.

Furthermore, \emph{unsupervised pre-training} of deep architectures~\cite{hinton06} can be regarded as forms of transfer learning, since the unsupervised learning task on which the model is trained in the first phase is different from the final, supervised task.


\subsection{Domain adaptation}\label{sec:domain-adaptation}
As previously stated, domain adaptation is a form of transfer learning where  out-of-domain data (i.e.~source domain) is used to improve the performance of a model solving the same task in the target domain. In this framework, the marginal distributions $P_s(Y|X)$  and $P_t(Y|X)$ are assumed to be neither identical nor independent~\cite{Daume:2006wj}. 
Previous work in domain adaptation use source data as ``prior knowledge'' to estimate the model parameters of the target domain using a Maximum a Posteriori (MAP) approach for tasks involving language modelling and parsing~\cite{Bacchiani:2003ii}. \citet{chelba2006adaptation} propose a model using a Maximum Entropy model for the MAP estimation of the model parameters of the target domain for capitalization of text. This strategy was later revisited in~\cite{daume07easyadapt}, where the parameters of a model trained in the source domain are used to regularise the parameters of a model trained in the target domain.


\subsection{Multi-task learning}\label{sec:multi-task-learn}
Multi-task learning is related to transfer learning in the sense that a model is used to address different (but related) tasks, but where most transfer learning problems involve a phase where a model is trained on one task before turning to another task, multi-task learning involves the simultaneous training of the model for the different tasks.
The motivation for this is that the training signals from related tasks serve to regularize the model, leading it to generalize better.
For example, in a task where the steering direction of a vehicle was to be predicted based on camera images of the road in front of the vehicle, \citet{caruana97:_multit_learn} found that it is beneficial to train the model (a neural network) simultaneously on a number of additional tasks, such as predicting the left and right borders of the road, and whether the road has one or two tracks.
Simultaneous training in this case was realized by having one output unit for each task, so effectively, all but the upper layer of the model are affected by the training on multiple tasks.


\subsection{Concept drift}\label{sec:concept-drift}
\emph{Concept drift} refers to the phenomenon where the target concepts of interest stay the same over time, but relation to inputs changes \cite{gama14:_survey_concep_drift_adapt}. 
As \cite{widmer96:_learn} note, models dealing with concept drift are required to hit a balance between stability and flexibility, since not all fluctuations in model performance may be attributable to concept drift. Noise may also account for temporal decreases in model performance.

\citet{Calandra:2012:LDB:2406821.2406873} propose a method to harness a deep belief network for non-stationary data streams. They intend to solve the problem of increased memory requirements when a model needs to be re-trained upon the arrival of data belonging to unseen labels. Rather than storing all old data, they create an artificial data set by generating data samples using the old model. Although this reduces the accuracy with respect to a model that is retrained on all available data, it has constant memory requirements.


\section{Method}\label{sec:method}

The specific problem we focus on in the current experiment is a form of transfer learning.
We are interested in the question how a model that has been trained to perform a particular task, can adapt to a novel task most effectively.
As illustrated in Figure~\ref{fig:transfer_plot}, there are several aspects to the notion of effective adaptation --- most importantly the asymptotic accuracy on the novel task, and the amount of training it takes to perform a task accurately.
We define the criteria we use for evaluation adaptation strategies in Section~\ref{sec:evaluation-criteria}, but before that, we will give a more precise description of the problem we are addressing.



\subsection{Problem description}\label{sec:problem-desc}

Given a labelled data set, we divide the data set into two subsets, such that one subset contains the data pertaining to one half of the labels, and the other subset contains the data pertaining to the other half of the labels.
We call one subset the \emph{source} domain, and the other the \emph{target} domain.
This implies that the label space is different, i.e.
$\Ycal_s \neq \Ycal_t$, and therefore $P_s(Y \mid X) \neq P_t(Y \mid X)$.
Although splitting the data set by labels does not strictly imply that the marginal data distributions of the domains are different, the labels usually relate to some morphological aspect of the features, so it is likely that  $P_s(X) \neq P_t(X)$.

The tasks we consider are classification, and autoencoding, and multi-task learning, in which a model is simultaneously trained to perform classification and autoencoding.
The classification task is formalized as $\Tcal^{CL} = \{ \Ycal, P(Y \mid X) \}$.
Note that in the autoencoding task, the labels $Y$ are not used.
Unfortunately, the probabilistic formalism is not well-suited to formalize the autoencoding task: autoencoding would be regarded as a case where $\Ycal = \Xcal$, which implies the trivial task $\Tcal^{AE} = \{ \Xcal, P(X \mid X) \}$.
The actual value of the autoencoding task is in the fact that the \emph{bottleneck} architecture of the models used to approximate $P(X \mid X)$ prevents them from learning a trivial mapping.
A more precise description of the autoencoding task is given in Section~\ref{sec:convae}.
Finally, multi-task learning involving simultaneous classification and autoencoding can be formalized as follows.
Instead of the label space \Ycal, we define $\Zcal = \Xcal \times \Ycal$.
The multi-task formalization is then $\Tcal^{MT} = \{ \Zcal,  P(Z | X) \} $, where $Z \in \Zcal$.

The aim of an adaptation strategy is then to transform a model that is optimal for task $\Tcal_s$ on $\Dcal_s$ into a model that is optimal on $\Tcal_t$ on $\Dcal_t$.
For example, a model trained for a classification task $\Tcal^{CL}$ in domain $\Dcal_s$, might be adapted to perform a classification task $\Tcal^{CL}$ in domain $\Dcal_t$, but it may also be adapted more radically, to perform autoencoding $\Tcal^{AE}$ in domain $\Dcal_t$.

Since the architecture of a model is determined by the task it performs, it is not obvious what we mean by adapting a model from one task to another.
In Section~\ref{sec:transfer-cnn-model} we will describe this procedure in more detail.
Another aspect to be clarified is how we measure the success of adaptation strategies.
We address this issue in Section~\ref{sec:evaluation-criteria}.






\subsection{Evaluation criteria}\label{sec:evaluation-criteria}
As described above, our goal is to find adaptation strategies for representation learning models that allow a model to perform as well as possible on a new task, with as little training as necessary.
In terms of the schematic plot in Figure~\ref{fig:transfer_plot}, our primary interest is in the slope of the learning curve, but obviously, a model that adapts quickly but has substantially lower asymptotic performance than is possible with a baseline method is undesirable.
Therefore, both the slope and the asymptotic performance with respect to some baseline methods should be taken into account when evaluating the methods.

We are ultimately interested in adaptation strategies that work well independent of the source and target tasks involved.
This suggests that we define a single quality measure, that aggregates the the slopes and asymptotic performances of a strategy over the different combinations of source and target tasks.
We chosen not to do so for the following reasons.
Firstly, since they are different quantities, there is no obvious way to combine the slope and asymptotic performance of an adaptation strategy in a principled manner.
A weighting of the two quantities necessarily reflects a personal judgement on their relative importance.
Secondly, the experiments reported here should be regarded as explorative.
Although we are interested in establishing the superiority of one adaptation strategy over another, independent of the tasks and domains involved, we have no a priori indication that such a ranking can be established across tasks and domains.

For these reasons, we refrain from defining a single evaluation criterion for the adaptation strategies.
We believe that more insight is gained by a qualitative analysis of the evolution of task performance in target domain for each of the adaptation strategies, as a function of training, using plots similar to the diagram in Figure~\ref{fig:transfer_plot}.

\subsection{Convolutional Neural Networks} \label{sec:cnn}
\begin{figure}
\begin{center}
\begin{tikzpicture}
\node at (0.75,-1){\tiny\begin{tabular}{c}Input \\ layer $l_0$\end{tabular}};

\draw[fill=black,opacity=0.1,draw=black] (0,0) -- (1.5,0) -- (1.5,1.5) -- (0,1.5) -- (0,0);

\draw (0.275,0.7) -- (0.475,0.7) -- (0.475,0.9) -- (0.275,0.9) -- (0.275,0.7);

\draw[dotted] (0.475,0.7) -- (2.7, 0.6);
\draw[dotted] (0.475,0.9) -- (2.7, 0.7);

\node at (3.0,-1){\tiny\begin{tabular}{c}Convolutional \\ layer $l_1$\end{tabular}};\draw[fill=black,opacity=0.1,draw=black] (3.3,0.8) -- (4.3,0.8) -- (4.3,1.8) -- (3.3,1.8) -- (3.3,0.8); 
\draw[fill=black,opacity=0.1,draw=black] (3.2,0.7) -- (4.2,0.7) -- (4.2,1.7) -- (3.2,1.7) -- (3.2,0.7); 
\draw[fill=black,opacity=0.1,draw=black] (3.1,0.6) -- (4.1,0.6) -- (4.1,1.6) -- (3.1,1.6) -- (3.1,0.6); 
\draw[fill=black,opacity=0.1,draw=black] (3.0,0.5) -- (4.0,0.5) -- (4.0,1.5) -- (3.0,1.5) -- (3.0,0.5); 
\draw[fill=black,opacity=0.1,draw=black] (2.9,0.4) -- (3.9,0.4) -- (3.9,1.4) -- (2.9,1.4) -- (2.9,0.4); 
\draw[fill=black,opacity=0.1,draw=black] (2.8,0.3) -- (3.8,0.3) -- (3.8,1.3) -- (2.8,1.3) -- (2.8,0.3); 
\draw[fill=black,opacity=0.1,draw=black] (2.7,0.2) -- (3.7,0.2) -- (3.7,1.2) -- (2.7,1.2) -- (2.7,0.2); 
\draw[fill=black,opacity=0.1,draw=black] (2.6,0.1) -- (3.6,0.1) -- (3.6,1.1) -- (2.6,1.1) -- (2.6,0.1); 
\draw[fill=black,opacity=0.1,draw=black] (2.5,0.0) -- (3.5,0.0) -- (3.5,1.0) -- (2.5,1.0) -- (2.5,0.0);

\draw (2.7,0.6) -- (2.8,0.6) -- (2.8,0.7) -- (2.7,0.7) -- (2.7,0.6);

\draw (3.15,0.1875) -- (3.35,0.1875) -- (3.35,0.3875) -- (3.15,0.3875) -- (3.15,0.1875);

\draw[dotted] (3.35,0.1875) -- (4.6375, 0.1875);
\draw[dotted] (3.35,0.3875) -- (4.6375, 0.2875);

\node at (4.875,-1){\tiny\begin{tabular}{c}Pooling \\ layer $l_2$\end{tabular}};\draw[fill=black,opacity=0.1,draw=black] (5.3,0.8) -- (6.05,0.8) -- (6.05,1.55) -- (5.3,1.55) -- (5.3,0.8); 
\draw[fill=black,opacity=0.1,draw=black] (5.2,0.7) -- (5.95,0.7) -- (5.95,1.45) -- (5.2,1.45) -- (5.2,0.7); 
\draw[fill=black,opacity=0.1,draw=black] (5.1,0.6) -- (5.85,0.6) -- (5.85,1.35) -- (5.1,1.35) -- (5.1,0.6); 
\draw[fill=black,opacity=0.1,draw=black] (5.0,0.5) -- (5.75,0.5) -- (5.75,1.25) -- (5.0,1.25) -- (5.0,0.5); 
\draw[fill=black,opacity=0.1,draw=black] (4.9,0.4) -- (5.65,0.4) -- (5.65,1.15) -- (4.9,1.15) -- (4.9,0.4); 
\draw[fill=black,opacity=0.1,draw=black] (4.8,0.3) -- (5.55,0.3) -- (5.55,1.05) -- (4.8,1.05) -- (4.8,0.3); 
\draw[fill=black,opacity=0.1,draw=black] (4.7,0.2) -- (5.45,0.2) -- (5.45,0.95) -- (4.7,0.95) -- (4.7,0.2); 
\draw[fill=black,opacity=0.1,draw=black] (4.6,0.1) -- (5.35,0.1) -- (5.35,0.85) -- (4.6,0.85) -- (4.6,0.1); 
\draw[fill=black,opacity=0.1,draw=black] (4.5,0.0) -- (5.25,0.0) -- (5.25,0.75) -- (4.5,0.75) -- (4.5,0.0);

\draw (5.7625,1.175) -- (5.9625,1.175) -- (5.9625,1.375) -- (5.7625,1.375) -- (5.7625,1.175); 
\draw (5.6625,1.075) -- (5.8625,1.075) -- (5.8625,1.275) -- (5.6625,1.275) -- (5.6625,1.075); 
\draw (5.5625,0.975) -- (5.7625,0.975) -- (5.7625,1.175) -- (5.5625,1.175) -- (5.5625,0.975); 
\draw (5.4625,0.875) -- (5.6625,0.875) -- (5.6625,1.075) -- (5.4625,1.075) -- (5.4625,0.875); 
\draw (5.3625,0.775) -- (5.5625,0.775) -- (5.5625,0.975) -- (5.3625,0.975) -- (5.3625,0.775); 
\draw (5.2625,0.675) -- (5.4625,0.675) -- (5.4625,0.875) -- (5.2625,0.875) -- (5.2625,0.675); 
\draw (5.1625,0.575) -- (5.3625,0.575) -- (5.3625,0.775) -- (5.1625,0.775) -- (5.1625,0.575); 
\draw (5.0625,0.475) -- (5.2625,0.475) -- (5.2625,0.675) -- (5.0625,0.675) -- (5.0625,0.475); 
\draw (4.9625,0.375) -- (5.1625,0.375) -- (5.1625,0.575) -- (4.9625,0.575) -- (4.9625,0.375);

\draw[dotted] (5.9625,1.175) -- (6.3875, 0.375);
\draw[dotted] (5.9625,1.375) -- (6.3875, 0.475);
\draw[dotted] (5.8625,1.075) -- (6.3875, 0.375);
\draw[dotted] (5.8625,1.275) -- (6.3875, 0.475);
\draw[dotted] (5.7625,0.975) -- (6.3875, 0.375);
\draw[dotted] (5.7625,1.175) -- (6.3875, 0.475);
\draw[dotted] (5.6625,0.875) -- (6.3875, 0.375);
\draw[dotted] (5.6625,1.075) -- (6.3875, 0.475);
\draw[dotted] (5.5625,0.775) -- (6.3875, 0.375);
\draw[dotted] (5.5625,0.975) -- (6.3875, 0.475);
\draw[dotted] (5.4625,0.675) -- (6.3875, 0.375);
\draw[dotted] (5.4625,0.875) -- (6.3875, 0.475);
\draw[dotted] (5.3625,0.575) -- (6.3875, 0.375);
\draw[dotted] (5.3625,0.775) -- (6.3875, 0.475);
\draw[dotted] (5.2625,0.475) -- (6.3875, 0.375);
\draw[dotted] (5.2625,0.675) -- (6.3875, 0.475);
\draw[dotted] (5.1625,0.375) -- (6.3875, 0.375);
\draw[dotted] (5.1625,0.575) -- (6.3875, 0.475);

\draw (4.6375,0.1875) -- (4.7375,0.1875) -- (4.7375,0.2875) -- (4.6375,0.2875) -- (4.6375,0.1875);

\node at (6.625,-1){\tiny\begin{tabular}{c}Convolutional\\ layer $l_3$\end{tabular}};\draw[fill=black,opacity=0.1,draw=black] (6.65,0.4) -- (7.4,0.4) -- (7.4,1.15) -- (6.65,1.15) -- (6.65,0.4); 
\draw[fill=black,opacity=0.1,draw=black] (6.55,0.3) -- (7.3,0.3) -- (7.3,1.05) -- (6.55,1.05) -- (6.55,0.3); 
\draw[fill=black,opacity=0.1,draw=black] (6.45,0.2) -- (7.2,0.2) -- (7.2,0.95) -- (6.45,0.95) -- (6.45,0.2); 
\draw[fill=black,opacity=0.1,draw=black] (6.35,0.1) -- (7.1,0.1) -- (7.1,0.85) -- (6.35,0.85) -- (6.35,0.1); 
\draw[fill=black,opacity=0.1,draw=black] (6.25,0.0) -- (7.0,0.0) -- (7.0,0.75) -- (6.25,0.75) -- (6.25,0.0);

\draw (6.3875,0.375) -- (6.4875,0.375) -- (6.4875,0.475) -- (6.3875,0.475) -- (6.3875,0.375);

\draw (6.7125,0.125) -- (6.9125,0.125) -- (6.9125,0.325) -- (6.7125,0.325) -- (6.7125,0.125);

\draw[dotted] (6.9125,0.125) -- (8.075, 0.125);
\draw[dotted] (6.9125,0.325) -- (8.075, 0.225);

\node at (8.25,-1){\tiny\begin{tabular}{c}Pooling\\ layer $l_4$\end{tabular}};\draw[fill=black,opacity=0.1,draw=black] (8.4,0.4) -- (8.9,0.4) -- (8.9,0.9) -- (8.4,0.9) -- (8.4,0.4); 
\draw[fill=black,opacity=0.1,draw=black] (8.3,0.3) -- (8.8,0.3) -- (8.8,0.8) -- (8.3,0.8) -- (8.3,0.3); 
\draw[fill=black,opacity=0.1,draw=black] (8.2,0.2) -- (8.7,0.2) -- (8.7,0.7) -- (8.2,0.7) -- (8.2,0.2); 
\draw[fill=black,opacity=0.1,draw=black] (8.1,0.1) -- (8.6,0.1) -- (8.6,0.6) -- (8.1,0.6) -- (8.1,0.1); 
\draw[fill=black,opacity=0.1,draw=black] (8.0,0.0) -- (8.5,0.0) -- (8.5,0.5) -- (8.0,0.5) -- (8.0,0.0);

\draw (8.075,0.125) -- (8.175,0.125) -- (8.175,0.225) -- (8.075,0.225) -- (8.075,0.125);

\draw[dotted] (8.5,0) -- (9.5, 0);
\draw[dotted] (8.5,0) -- (10.7374368671, 1.23743686708);
\draw[dotted] (8.9,0.9) -- (9.5, 0);
\draw[dotted] (8.9,0.9) -- (10.7374368671, 1.23743686708);

\node at (10.0,-1){\tiny\begin{tabular}{c}Dense \\ layer $l_5$\end{tabular}};\draw[fill=black,draw=black,opacity=0.5] (9.5,0) -- (10.0,0) -- (11.2374368671,1.23743686708) -- (10.7374368671,1.23743686708) -- (9.5,0);

\node at (11.75,-1){\tiny\begin{tabular}{c}Dense \\ layer $l_6$\end{tabular}};\draw[fill=black,draw=black,opacity=0.5] (11.5,0.5) -- (12.0,0.5) -- (12.5303300859,1.03033008589) -- (12.0303300859,1.03033008589) -- (11.5,0.5);

\draw[dotted] (10.0,0) -- (11.5, 0.5);
\draw[dotted] (10.0,0) -- (12.0303300859, 1.03033008589);
\draw[dotted] (11.2374368671,1.23743686708) -- (12.0303300859, 1.03033008589);
\draw[dotted] (11.2374368671,1.23743686708) -- (11.5, 0.5);
\end{tikzpicture}
\caption{Example of the architecture of a Convolutional Neural Network classifier, including two convolutional-pooling substages (layers $l_1$ and $l_2$, and $l_3$ and $l_4$), and a classification stage consisting of two fully connected layer (as a coding substage $l_5$, and the proper classifier $l_6$).} \label{fig:cnn_architecture}
\end{center}
\end{figure}
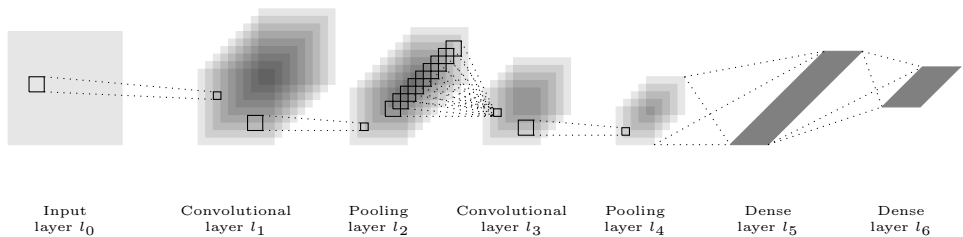

Convolutional Neural Networks (CNNs) are a special kind of Feed Forward Neural Networks (FFNNs) that build some invariance properties into the structure of the neural network~\cite{Bishop:2006}.
CNNs have been successfully used in several machine learning applications, including natural language processing and image classification~\cite{Krizhevsky:2012wl}.
CNNs have a number of advantages over fully connect FFNNs. Firstly, the convolutional nature of the architecture, using small convolutional filters enforces the extraction of local features~\cite{Lecun:1998hy}.
Secondly, they typically have shared weights, which greatly reduces number of parameters compared with similar sized FFNNs~\cite{Krizhevsky:2012wl}.
Lastly, they typically perform spatial sub-sampling, which adds robustness against noise and local distortions.
A typical CNN has three building blocks: convolutional, subsampling, and dense (fully connected) layers.
A CNN is illustrated in Figure \ref{fig:cnn_architecture}.

The basic building block of CNNs are the convolutional layers.
The input of $l$-th convolutional layer consists of $m_1^{(l-1)}$ feature maps from the previous layer, while its output consists of $m_1^{(l)}$ feature maps.
The $i$-th feature map in this layer is given by
\begin{equation}
\mathbf{y}_i^{(l)} = f_l\left(\sum_{j=1}^{m_1^{(l-1)}} \mathbf{W}^{(l)}_{i,j}*\mathbf{y}_{j}^{(l-1)}  + \mathbf{B}_{i}^{(l)}\right),
\end{equation}
where $*$ represents the convolution of $\mathbf{y}_{j}^{(l-1)}$ with $\mathbf{W}_{i,j}^{(l)}$, a kernel of size $h_1^{(l)}  \times h_2^{(l)}$ connecting the $j$-th feature map in layer $(l-1)$ with the $i$-th feature map in layer $l$, $\mathbf{B}_i^{(l)}$ is a bias (matrix) and $f_l$ is an elementwise non-linear activation function.

Pooling can be understood as a form of non-linear downsampling.
Max-Pooling layers partition an input feature map into a set of non-overlapping rectangles (\emph{pool}), and for each such pool, outputs its maximum value.

Finally, dense layers are the standard fully connected layers in FFNNs.
The output of the $l$-th dense layer can be computed as
\begin{equation}
\mathbf{y}^{(l)} = f_l\left(\mathbf{W}^{(l)} \mathbf{y}^{(l-1)} + \mathbf{b}^{(l)} \right),\label{eq:fully_connected}
\end{equation}
where $\mathbf{W}^{(l)}$ is a filter connecting layer $(l-1)$ to layer $l$, $\mathbf{b}^{(l)}$ is a bias vector and $f_l$ is an elementwise non-linear activation function.
The set of all parameters of a CNN, i.e.\ kernels, filters and biases, will be denoted \params.

Common activation functions for CNNs include sigmoid, $\tanh$,  rectifier, and softmax, which is particularly useful as the output of a multi-class classifier~\cite{Bishop:2006}. See Appendix \ref{sec:act_functions} for explicit definitions of these activation functions.

In practice, convolutional and pooling layers are used to learn a feature hierarchy, while dense layers are used for classification purposes based on the computed features \cite{Lecun:1998hy}.
In the following, we will refer to the stack of a convolutional and pooling layers as a \emph{convolutional stage}, and the stack of dense layers as fully connected stage, or as a classification stage, if its primary objective is classification.

The classification task for CNNs can be formally described as follows.
Given a set of input images $\{\mathbf{x}_1, \dots, \mathbf{x}_N\}$, and a set of targets $\{ \mathbf{t}_1, \dots, \mathbf{t}_N\}$, where $\mathbf{t}_i$ is a one-hot encoding of the class of $\mathbf{x}_i$, the parameters of a CNN can be learned in a supervised way as
\begin{equation}
\hat{\params} = \argmin_{\params} L (\params),\label{eq:parameter_learning}
\end{equation}
\noindent where $L(\params)$ is the loss function.
The standard loss function for a multi-class classification problem is the  mean categorical cross entropy\footnote{See Eq.~\eqref{eq:CCE} in Appendix \ref{sec:cost_functions}.}~\cite{Bishop:2006}.

\subsubsection{Convolutional autoencoders} \label{sec:convae}
Most methods in unsupervised learning are based on the ``encoder-decoder'' paradigm~\cite{Masci:2011tz}, where the input is first transformed into a (typically) lower-dimensional space (\emph{encoding stage}) and then expanded to reproduce the initial data (\emph{decoding stage}).
Examples of this paradigm include Low-Complexity Coding and Decoding Machines, Predictability Minimization layers, Restricted Boltzmann Machines and autoencoders~\cite{Masci:2011tz}.

An autoencoder (AE) is a particular neural network architecture used for feature learning.
Its aim is to learn an encodings, i.e.,  distributed (and usually compact) representation of a set of data.
 Formally, an AE takes an input $\fatx \in \mathbb{R}^{N_x}$ and maps it to a latent  representation or \emph{encoding} $\mathbf{y}_h \in \mathbb{R}^{N_h}$, using a deterministic function of type
\begin{equation}
\mathbf{y}_h = f\left(\fatx ; \params \right) = f\left( \mathbf{W} \fatx + \mathbf{b} \right)
\end{equation}
where $\params=\{ \mathbf{W}, \mathbf{b}\}$ are the encoding parameters.
A typical autoencoder uses functions similar to those of the fully connected layers (see Eq.~\eqref{eq:fully_connected}).
This code is used to reconstruct the input by a reverse mapping (\emph{decoding}), i.e.,
\begin{equation}
\mathbf{y}_{decoding} = f\left(\mathbf{y}_h; \params^{'} \right) = f\left(\mathbf{W}^{'} \mathbf{y}_h + \mathbf{b}' \right),\label{eq:decoding_fully_connected}
\end{equation}
where $\params^{'}=\{\mathbf{W}^{'}, \mathbf{b}^{'}\}$ are the decoding parameters.
A usual constraint of the decoding parameters if for the weights to take the form $\mathbf{W}^{'} = \mathbf{W}^T$, i.e., use the same weights for encoding of the input and decoding of the latent representation.

Convolutional Autoencoders (CAEs) are (deep) autoencoders that use CNNs in the in the encoding stage~\cite{Masci:2011tz}.
As previously stated in Section \ref{sec:cnn}, CNNs allow for discovery of localized features that appear over 2D input.
In a CAE, the reconstruction of the input data is due to a combination of basic image patches based on the latent code.

The decoding layers corresponding to fully connected layers are given in the same form as in Eq.~\eqref{eq:decoding_fully_connected}.
For convolutional layers, the corresponding $i$-th feature map of decoding layer $l$ is given by
\begin{equation}
\mathbf{y}^{(l)}_i = f\left(\faty^{h_{(l-1)}}; \params^{'} \right) = f\left(\sum_{j=1}^{m_l^{h_{(l-1)}}} \tilde{\mathbf{W}}_{i,j}^{(l)}  * \mathbf{y}_j^{h_{(l-1)}} + \mathbf{B}'^{(l)}\right),
\label{eq:decoding_convolutional}
\end{equation}
where $\faty^{h_{(l-1)}}$ represents all $m_l^{h_{(l-1)}}$ feature maps of the encoding layer  $h_{(l-1)}$, $\tilde{\mathbf{W}}$ represents the flip operation over both weight dimensions of the corresponding encoding parameter $\mathbf{W}$, and $\mathbf{B}'^{(l)}$ is a bias.

For downsampling layers in the encoding stage, the corresponding decoding layer corresponds to an upsampling layer.
One possible upsampling strategy for max-pooling layers  is to ``remember" the position of the input that had the maximum value for every max-pooling during the forward propagation of the input data through the network.
We will refer to a a layer that implements such a strategy as an \emph{unpooling} layer.

Given a set of training data $\mathbf{X} =\{\mathbf{x}_1, \dots, \fatx_N\}$, the parameters of a (C)AE can be optimized in an unsupervised way by minimizing the reconstruction error of $\mathbf{X}$, in a similar fashion to Eq.~\eqref{eq:parameter_learning}.
The typical loss function for (C)AEs is the mean squared error\footnote{See Eq.~\eqref{eq:MSE} in Appendix \ref{sec:cost_functions}.}.

\subsubsection{Multi-task learning}\label{sec:multitask}
Solving a problem constrained to multiple objective functions is the object of study of \emph{muticriterion optimization}~\cite{Boyd:2004uz}.
A vector optimization problem can be formalized as follows.
Given a vector loss function $\mathbf{L}(\params) = [L_1,\dots,L_q]^T$, whose components can be interpreted as $q$ different scalar objectives, to be minimized.
In order to apply standard scalar optimization methods, such as gradient descent, we can ``scalarise'' a multi-criterion problem by forming a weighted sum objective, i.e.
\begin{equation}
L = \boldsymbol\alpha^T \mathbf{L}(\params) = \sum_{i=1}^q \alpha_i L_i(\params),
\end{equation}
where $\alpha_i$ represents a weighting coefficient for the $i$-th component of $\mathbf{L}(\params)$.
This weights are usually constrained to $\alpha \leq 0$ and $\sum_i^q\alpha_i = 1$ .
Using this framework, it is possible to express the multi-task objective as minimizing a joint loss function for both classification and autoencoding as follows
\begin{equation}
L(\params) = (1 - \alpha_{MT}) L_{CL}(\params) + \alpha_{MT} L_{AE}(\params),
\end{equation}
where $\alpha_{MT}$ is the multi-task weight coefficient and $L_{CL} = L_{CCE}$ and $L_{AE} = L_{MSE}$ represent the loss functions for the classification and autoencoding tasks, respectively.

\subsubsection{Transfer of a CNN model from one task to another}\label{sec:transfer-cnn-model}
Given a CNN  with $N$ layers used for classification, an CAE can be built as follows.
For layers $l_{N-1}, \dots l_{1}$, (i.e., all layers in the CNN except the classification stage $l_N$), build the mirroring decoding layer, corresponding to Eq.~\eqref{eq:decoding_fully_connected} if the encoding layer is a fully connected layer, Eq.~\eqref{eq:decoding_convolutional} if the encoding layer is a convolutional layer or an unpooling (upsampling) layer if the encoding layer is a max-pooling (subsampling) layer.
Conversely, given a CAE with $2(N -1)$ layers, a CNN classifier can be built by removing the decoding layers from the CAE (layers $l_N$ to $ l_{2(N-1)}$) and appending a fully connected layer with with as many softmax units as the desired number of classes.

In order to reduce the degrees of freedom, in the transfer of a model from one task to another, we do not use biases in the decoding layers of a CAE built from an autoencoder.
This guarantees that only the weights (present in both models) are responsible for learning the task.

\subsection{Strategies for conceptual change in CNNs}\label{sec:strat-conc-change}
In this Subsection, we propose a number of strategies for adapting learned representations in response to new tasks.
For clarity, we label each choice with a keyword, and refer to the final strategies as a combination of keywords, listed at the end of this section.

\subsubsection{Two baselines}\label{sec:two-baselines}
There are two obvious baseline approaches to the adaptation of a model to a new task.
One is to make no changes to the model upon the change of task.
In this case, the only change is that the training data of the old task is replaced by the training data of the new task.

A practical issue is that a new task implies a new interpretation of the outputs, and possibly a different number of outputs.
Reusing the parameters of the output layer for the new task raises the question of how the output variables for the old task should be mapped to those of the new task, a mapping that is necessarily arbitrary.
For this reason, when we reuse the parameters of a prior model, we replace its output layer (the rightmost layer in Figure~\ref{fig:cnn_architecture}, page~\pageref{fig:cnn_architecture}) with a new layer, initialized with random parameter values.
Ignoring this detail, we refer to this baseline strategy as REUSE ALL.

Another, contrary approach is to altogether ignore the representations learned on the prior task, and start with a randomly initialized model to learn the new task.
This can be regarded as a case of infinite plasticity, where the new task forms the representations without any trace of the representations that were learned in the prior task.
This strategy will be referred to as RESET.

\subsubsection{Selective reuse of the prior model: keep convolutional filters}\label{sec:select-reuse-conv-filt}
Considering the dynamics of learning, REUSE ALL may not be ideal, since it starts learning task 2 with a model that is specialized on task 1.
It may take more effort to ``unlearn'' aspects of task1 that are irrelevant for task 2, than to learn task 2 from scratch.

However, if the data in task 1 and task 2 are similar in nature, it is likely that the data have at least some common structure.
For example, in the case of natural images depicting different classes of objects, it is likely that certain low level representations, such as local edges in natural images, are useful for different tasks, such as the recognition of different object classes.
Different classes of objects may involve different constellations of similar forms.
For example, a dark round shape may represent wheels on a car, portholes in a ship, or the eyes of an animal.

In the current experiment, this observation inspires an adaptation strategy where the convolutional filters --- representing the low level representations --- are preserved across tasks, whereas the rest of the network is re-initialized to a random state.
This option is referred to as REUSE CF.



\subsubsection{Prior regularisation}\label{sec:prior-regularisation}

A strategy for domain adaptation proposed by \citet{chelba2006adaptation} is to take a maximum a-posteriori approach to the estimation of the model parameters for the new task, where the model learned on the first task provides a prior estimate of the parameters.
This prior serves as a regulariser for the parameters.

Standard parameter regularisation schemes are based on the assumption that the parameters \params{} follow a zero-mean gaussian distribution, that is $p(\params) \sim \mathcal{N}(\mathbf{0}, \frac{1}{\lambda} \mathbf{I})$, leading to the addition of a regularisation term $\frac{\lambda}{2} ||\params||^2_2$ to the standard expression for the loss, in Equations~(\ref{eq:MSE}) and (\ref{eq:CCE}), page~(\pageref{eq:MSE}).

The prior proposed by \citet{chelba2006adaptation} is:
\begin{equation}
\label{eq:prior}
p(\params) \sim \mathcal{N}(\oldparams, \frac{1}{\lambda} \mathbf{I}),
\end{equation}
a gaussian distribution around the parameters \oldparams{} that were learned on the first task.
This assumption leads to an alternative loss function $L_\mathit{PR}$, including the prior regularisation ($\mathit{PR}$) term \cite{daume07easyadapt}:

\begin{equation}
\label{eq:priorreg}
L_\mathit{PR}(\params) = L(\params) + \frac{\lambda}{2} ||\oldparams - \params||^2_2.
\end{equation}

In the current experiment, the prior regularisation is applied to the convolutional filters, and used in combination with the RESET option; it is referred to as RESET PRF.

\subsubsection{A note on random initialization}\label{sec:note-rand-init}
The idea of investigating the variance of the layer outputs to improve weight initialization for deep learning was introduced in~\cite{Glorot:2010uc}.
This result has motivated the search for careful initialization, rather than unsupervised pretraining of networks with methods such as RBMs, thus representing a considerable speedup in the training of neural networks .
\citet{Glorot:2010uc} suggest  that keeping the layer-to-layer transformations such that the singular values of the Jacobian matrices associated with each layer\footnote{the Jacobian matrix associated with the $l$-th layer is given by $\mathbf{J}^{(l)}= \dpar{\mathbf{y}^{(l+1)}}{\mathbf{y}^{(l)}}$.} are approximately  1 is equivalent to keeping the ratio of the average activation variance going from layer $\mathbf{y}^{(l)}$ to layer $\mathbf{y}^{(l+1)}$.
 This result implies that the random initialization of the model parameters for each layer depends on the nonlinear activation function and the size of the layer.
The main intuition behind this initialization strategy is that if the network parameters are initialized to small, the output shrinks while passing through each layer, and eventually being to small.
On the other hand, if the network parameters are too large, the output of each layer keeps growing, until the output of the network is saturated.

We use the same random initialized parameters across adaptation strategies to rule out initialization as a noise factor for each run of the experiment.


\section{Experiment}\label{sec:experiment}
The adaptation strategies are compared on three different data sets, described in Section~\ref{sec:datasets}.
For each data set the following procedure is followed.

\begin{enumerate}
\item The data set is split into a source and a target domain, as described in Section~\ref{sec:problem-desc}.
For the specific partitioning per data set, see Section~\ref{sec:datasets}
\item A model is trained in the source domain for each of three tasks: classification, autoencoding, and the multi-task of simultaneous classification and autoencoding
\item A model is trained in the target domain for each of two tasks: classification, and autoencoding.
For each adaptation strategy, two training runs are realized
\item Each model is evaluated on the target test set during training, in order to monitor adaptation.
Per adaptation strategy, the results of the two training runs are averaged, in order to reduce the impact of random effects
\end{enumerate}

\begin{figure}[ht!]
\centering

\includegraphics[width=\linewidth]{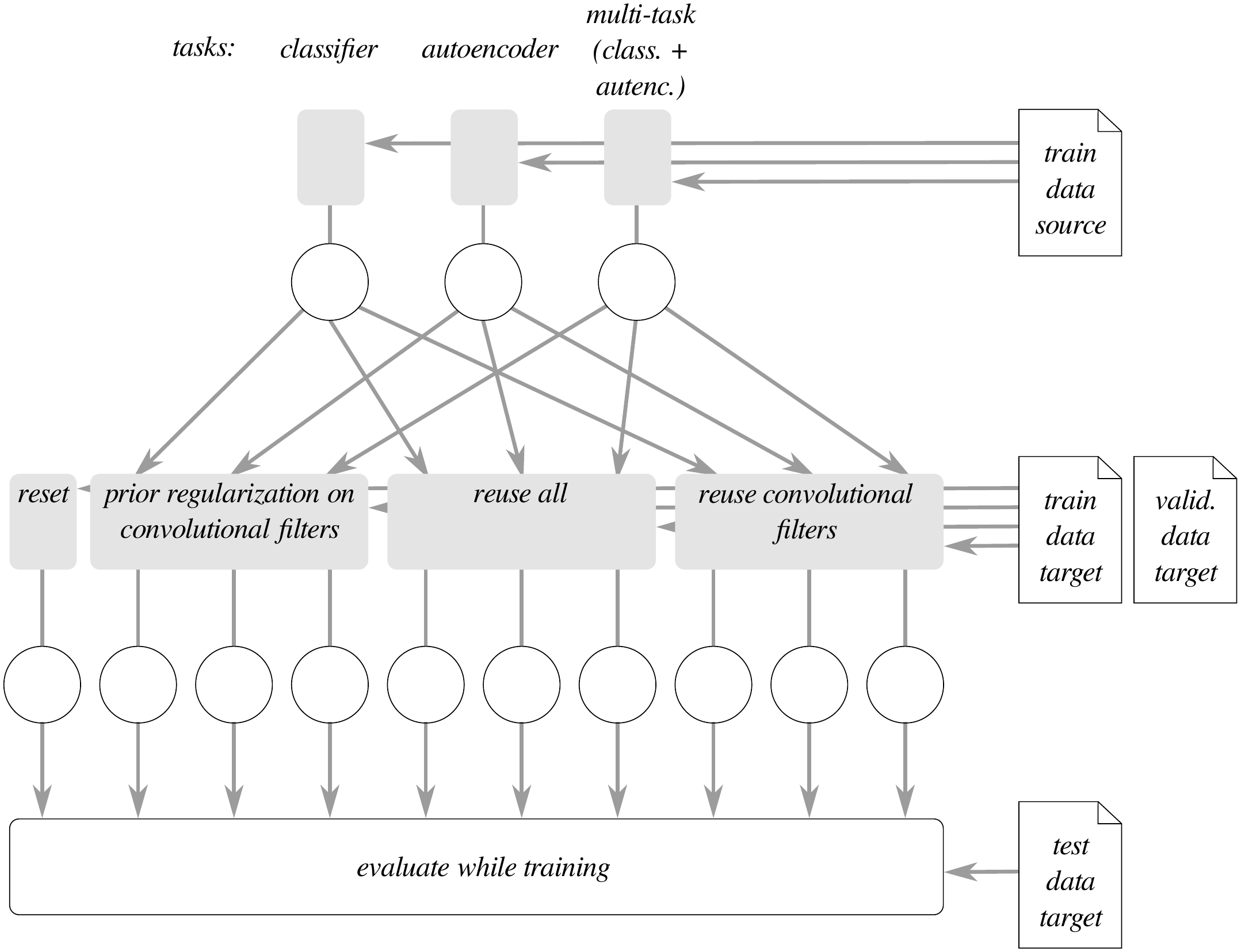}
\caption{Schematic overview of the experiment for a single data set; grey rounded boxes represent training methods, circles represent models, document shapes represent data instances, and the white rounded box represents the evaluation method}
\label{fig:diagram}
\end{figure}

A schematic overview of the above process is given in Figure~\ref{fig:diagram}.
Since the above process produces a multitude of models, each trained under different conditions, we will refer to them using combinations of labels denoting the adaptation strategy used, and labels denoting the prior model used by the adaptation strategy.
The labels are listed in Table~\ref{tab:adapt-strategies}.

\begin{table}[t]
\centering
\begin{tabular}[t]{lp{.6\linewidth}}
RESET & Initialize parameters with random values \\
RESET PRF & Prior regularisation on convolutional filters \\
REUSE ALL & Initialize all parameters (except output layer) from prior model \\
REUSE CF & Initialize only convolutional filter parameters from prior model \\
(CL) & Prior model trained as classifier \\
(AE) & Prior model trained as autoencoder \\
(MT) & Multi-task: Prior model trained simultaneously as classifier and autoencoder
\end{tabular}
\caption{Labels denoting adaptation strategies, and their meaning, as used in the results.
The labels in parentheses represent prior models, to be used in conjunction with one of the adaptation strategies}
\label{tab:adapt-strategies}
\end{table}

Since we are interested in adaptation methods that allow for a quick adaptation of a model from the source task/domain to the target task/domain, we intentionally limit the size of of the training (and validation) set in the target domain.
The scarcity of training samples makes it harder for the model to adequately generalize in the target domain, and thus it increases the potential benefit of adapting a model from another domain (although the actual benefit of course depends on the resemblance of the domains).


\subsection{Data Sets}\label{sec:datasets}
\paragraph{MNIST} \label{sec:datasets:mnist}
The Mixed National Institute of Standards and Technology (MNIST) database  consists of handwritten digits collected by American high school students and employees of the United States Census Bureau~\cite{Lecun:1998hy}.
This database constitutes one of the most used data sets for benchmarking machine learning algorithms~\cite{Bishop:2006}.
The MNIST consists of 70,000  gray-scaled images, rescaled to fit in a $20\times20$ pixel box, and then centred in a $28\times20$ box.
For these experiments, MNIST was divided into two subsets, namely Data Set 1, consisting of digits $\{0, 1, 2, 3, 4\}$, and Data Set 2, consisting of digits $\{5, 6, 7, 8, 9\}$.
The training set is divided into 50,000 examples for computing the parameter updates, 10,000 examples for validation and 10,000 examples for testing.
Data Set 1 contains 25,538 examples for training, 5,058 for validating and 5139 for testing.
For Data Set 2, 10 samples per class were randomly selected for both training and validating, making a total of 50 examples for training, 50 for validating and 4861 for testing.

\paragraph{CIFAR-10}\label{sec:datasets:cifar10}
The CIFAR-10 is a labelled subset of the \emph{80 million tiny images} data set collected by Krizhevsky, Nair and Hinton~\cite{Krizhevsky:2009tr}.
This data set has been used to evaluate the performance of algorithms in machine learning and computer vision.
CIFAR-10 consists of 60,000 $32\times 32$ colour (RGB) images divided into 10 classes, that comprising vehicles and animals.
In this paper, Dataset 1 was chosen to include all instances of classes ``airplane'', ``automobile'', ``bird'', ``cat'', and ``deer'', while Data Set 2 consists  of classes ``dog'', ``frog'', ``horse'', ``ship'', and ``truck''.
Each training set was split into training and a validation sets, consisting in $75\%$ (37,500 examples) and $25\%$ (12,500 examples) of the data, respectively, while the test set contains 10,000 examples. Data Set 1 consists of 18,681, 6,319 and 5,000 examples, respectively for training, validating and testing, while  Data Set 2 contains 10 randomly selected samples per class for training and 10 randomly selected samples per class for testing, to make to make a total of 50 training examples, 50 validating examples and 5,000 examples for testing.
\paragraph{Composers}\label{sec:datasets:composers}
As an application of the proposed methods in a musical domain, a data set consisting of excerpts of musical pieces of the baroque and classical periods was used.
These excerpts are represented as piano-rolls, i.e., images where each pixel in the y axis corresponds to a musical note (using the MIDI note number convention), and each pixel in the x axis represents a unit of time.
 The scores where taken from Muse Data database\footnote{\url{http://www.musedata.org}}, an electronic library of classical music scores created by the Center for Computer Assisted Research in the Humanities at Stanford University.
Each excerpt has a length of 50 quarter notes, with a sample rate of a 32nd note (an 8th of a quarter), with a hop size of 10 quarter notes between contiguous windows.
All pieces have been centred to fit in a MIDI range of 68 notes.
To signalize the end of a note, the its last 32nd is left blank.
Data Set 1 consists of a selection from excerpts from G.
P.
Telemann's Cantatas, J.
S.
Bach's Cantatas and G.
F.
H\"andel's Concerti Grossi and Trio Sonatas,  while Data Set 2 consists of excepts from string quartets by F.
J.
Haydn and W.
A.
Mozart.
All excerpts coming from a single piece\footnote{Movements of a work are considered individual pieces.} appear only in either the training, validation or test sets, i.e., there are no pieces that appear on more than one  data set.
There are a total of 2073 training examples, 1393 validation examples and 993 testing examples, with Data Set 1 containing 1141, 755 and 541 examples for training, validating and testing, respectively and Data Set 2 consisting of 10 randomly selected samples per class for training and 10 randomly selected samples per class for validation, to make a total of 20 examples for training, 20 examples for validating and 452 examples for testing.

\subsection{Model training}\label{sec:model_training}

All models were trained using RMSProp~\cite{Dauphin:2015wn}, a derivative of the traditional backpropagation algorithm.
This method is a mini batch variant of stochastic gradient descent that adaptively adjusts the learning rate by dividing the gradient by an average of its recent magnitude.
In order to accelerate gradient descent, we use Nesterov's method for accelerating gradient descent~\cite{Sutskever:2013uwa}.
 In order to avoid overfitting, several strategies are used, including $l_2$-norm weight regularisation, enforcing sparseness in layer activations, early stopping  and dropout.
Regularisation of the $l_2$ norm enforces sparse parameters~\cite{Bishop:2006}, while the sparsity in layer activations was enforced using Hoyer's sparseness measure~\cite{2004cs........8058H}.
Dropout prevents overfitting and provides a way of approximately combining different neural networks efficiently by randomly removing units in the network, along with all its incoming and outgoing connections~\cite{Srivastava:2014ww,Hinton:2012wr}.
The network architectures and hyper-parameters were empirically selected by optimizing the models to their respective validation sets.

\paragraph{MNIST}\label{sec:training:mnist} The classifier consists of a CNN with a convolutional layer with 32 kernels of size $5 \times 5$ and sigmoid activations ($l_1$), followed by a max-pooling layer with a pool size of $2\times2$ ($l_2$).
The convolutional stage is followed by a classification stage consisting of a fully connected layer with 40 sigmoid units and ($l_3$) and a fully connected layer with 10 softmax units.
Dropout is used after the convolutional stage, i.e., after $l_2$.

The learning rate for RMSProp is set to $10^{-5}$, Nesterov's momentum to 0.5, the probability of dropout is set to 0.5, the regularisation coefficient is $0.001$, the sparsity coefficient is set to $10^{-4}$, the target sparseness is $0.9$ and the batch size is 50.
 The network was trained for a maximum of 2000 epochs, with a maximum of 200 epochs from the best result for early stopping.
The multi-criterion weighting coefficient is set to $0.01$ in multi-task models.

\paragraph{CIFAR-10} The classifier consists of a CNN with a convolutional layer with 32 kernels of size $5 \times 5$ and sigmoid activations ($l_1$), followed by a max-pooling layer with a pool size of $2\times2$ ($l_2$).
The classification stage consists of a fully connected layer with 40 sigmoid units and ($l_3$) followed by a fully connected layer with 10 softmax units. As in the previous case, dropout is used after the convolutional stage, i.e., after $l_2$.

The learning rate for RMSProp is set to $10^{-5}$, Nesterov's momentum to 0.5, the probability of dropout is set to 0.5, the regularisation coefficient is $0.001$, the sparsity coefficient is set to $10^{-4}$, the target sparseness is $0.9$ and the batch size is 50.
 The network was trained for a maximum of 3000 epochs, with a maximum of 200 epochs from the best result for early stopping.
In multi-task models, the multi-criterion weighting coefficient is set to $0.01$.

\paragraph{Composers}  The classifier consists of a CNN with a convolutional layer with 9 kernels of size $9 \times 9$ and rectified linear activations ($l_1$), followed by a max-pooling layer with a pool size of $2\times2$ ($l_2$), then a second convolutional layer with 5 kernels of size $5\times 5$ with rectified linear activations ($l_3$), followed by a max-pooling layer with a pool size of $2\times 2$ ($l_4$).
The convolutional stage is followed by a fully connected layer with 256 rectified linear units and ($l_5$) and fully connected layer with 4 softmax units ($l_6$), both conforming the classification stage. In a similar fashion as in the previous cases, dropout is used after the convolutional stage, i.e., after $l_4$.

The learning rate for RMSProp is set to $10^{-6}$, Nesterov's momentum to 0.5, the probability of dropout is set to 0.5, the regularisation coefficient is $0.001$, the sparsity coefficient is set to $10^{-4}$, the target sparseness is $0.5$ and the batch size is 50.
 All networks was trained for a maximum of 2000 epochs, with a maximum of 200 epochs from the best result for early stopping.
For the models solving the joint classifier and autoencoding task, the multi-criterion weighting coefficient is set to $0.01$.







\section{Results and discussion}
\label{sec:resanddisc}
The results of the various adaptation scenarios and data sets are displayed in Figures~\ref{fig:acc},~\ref{fig:clloss}, and ~\ref{fig:aeloss}.
Figure~\ref{fig:acc} shows the classification accuracies for the classification task $\Tcal_t^{CL}$ in the target domain $\Dcal_t$, for each of the three data sets.
The fact that the accuracy curves are rather noisy is because the classification accuracy is not the training objective, but rather the categorical cross-entropy of the model output with the one-hot representation of the class labels (plotted in Figure~\ref{fig:clloss}).
Furthermore, the occasional discontinuities in the curves are due to the averaging over multiple runs, where some runs converge (and thus halt) after fewer epochs than others.

The first baseline strategy (RESET) consists in a random re-initialisation of all parameters (conform Section~\ref{sec:note-rand-init}), meaning that no knowledge from the source task $\Tcal_s$ in the source domain $\Dcal_t$, is used at all.
The second baseline strategy (REUSE ALL), is to initialize the model with the parameters of the model trained on $\Tcal_s$ in the source domain $\Dcal_t$.
Note that there are three source tasks (CL, AE, and MT).

In Figure~\ref{fig:clloss}, there is a consistent trend that the \mbox{REUSE ALL} baseline strategy is slow to adapt, independent of the source task.
Interestingly, among the \mbox{REUSE ALL} conditions (dotted lines), the adaptation of the autoencoder model of the source domain---\mbox{REUSE ALL (AE)}---is more beneficial to learning the target classification task than the adaptation of the classifier---\mbox{REUSE ALL (CL)}---from the source domain.
This may be an indication that the autoencoder learns representations that are useful for encoding the data in general, whereas the classifier learns more specialised features that are useful primarily for recognising the specific classes that happen to be in the source domain.
To some degree, this finding underlines the rationale for \emph{unsupervised pre-training}~\cite{Erhan:2010wd}, that learning to encode the data independent of any classification objective provides more robust representations (for subsequent classifcation) than driving the representations by a classification objective (with different class labels) from the start.
The fact that the unsupervised-pretraining---\mbox{REUSE ALL (AE)}---does not surpass the \mbox{RESET} strategy suggests that the distribution of the data that the autoencoder has seen is not sufficiently representative for the target domain, i.e.
$P_s(X) \neq P_t(X)$.

The low loss values of \mbox{REUSE ALL (AE)} with respect to \mbox{REUSE ALL (CL)} and \mbox{REUSE ALL (MT)} does translate into higher accuracy rates (Figure~\ref{fig:acc}) for the CIFAR and Composers data sets, but surprisingly, it does not for the MNIST data set.

The \mbox{REUSE CF} strategy (Figure~\ref{fig:clloss}, dashed lines), which retains the first layer of convolutional filters learnt from the source domain, but resets all other parameters of the model, considerably speeds up adaptation to the target task with respect to the \mbox{RESET} baseline.
With this strategy, the adaptation from the classification task in the source domain is generally most successful, but even when adapting models from the AE and MT source tasks, the adaptation is mostly quicker than \mbox{RESET}.
A plausible explanation for this is that the low level structure of the data is usually more general, whereas the higher level structure (say, the configuration of lower level features into shapes) is more specific to particular data categories.

For the autoencoding task $\Tcal_t^{AE}$ in the target domain $\Dcal_t$ (Figure~\ref{fig:aeloss}), the \mbox{REUSE ALL (AE)} strategy leads to substantially better autoencoding results from the start in the target domain, but as opposed to the other adaptation strategies, the training of the model using this strategy does hardly improve the autoencoding loss beyond its initial value.
The \mbox{REUSE ALL (CL)} strategy on the other hand, adapts much worse on than the \mbox{RESET} baseline on CIFAR and MNIST, on a par with \mbox{REUSE CF (CL)}.

The autoencoder task in the target domain of the Composers data set shows a very different pattern (Figure~\ref{fig:aeloss}, right).
Here, all adaptation strategies outperform the \mbox{RESET} baseline that discards any knowledge from the source domain.
This may be an indication that the source and target domains in the Composers data set have rather similar marginal distributions, i.e., $P_s(X) \approx P_t(X)$, giving the adaptation strategies an advantage over the \mbox{RESET} baseline.
An explanation for this may be that there is a lot of structure in the musical data that is common across the music of the different composers (tonal structure, rhythmical structure).
The prevalence of this common structure would also explain the rather low overall accuracy in the composer classification task (Figure~\ref{fig:acc}, right).

The simultaneous learning of both autoencoding and classification in the source domain appears to provide more useful knowledge to be transferred to the autoencoding target task (Figure~\ref{fig:aeloss}), than the classification target task (Figures~\ref{fig:acc}, and ~\ref{fig:clloss}).
In the latter case, adaptation is more successful when the source task is also classification.

In the prior regularization strategy (\mbox{RESET PRF}), the first layer convolutional filters of a randomly initialised network are biased towards the filters learned from the source domain and task.
This regularization does improve a bit upon the \mbox{RESET} baseline sometimes, but not very consistently, and the gains are usually moderate.

One issue that we have not discussed is whether the asymptotic performance of some strategies is substantially different from that of others.
From the Figures it is clear that broadly speaking, the adaptation strategies converge to a certain range, even if there are individual differences.
To make better judgements on the asymptotic performance of the adaptation strategies however, it is necessary to do more extensive experimentation, with longer training phases, more data sets, and averaging results from more runs of the same condition.
Moreover, although the hyperparameters of the models (e.g.
the number of convolutional filters, the pooling size, regularisation constants) were selected through exploration prior to the experiment, a more exhaustive study of the hyperparameters may provide a better view on the value of the adaptation strategies.
From a computational point of view, this is a considerable undertaking, given that the results presented here require several weeks of continuous computation on multiple GPGPU-enabled machines.

\begin{landscape}\centering
\begin{figure}
\includegraphics[height=1.15\textheight]{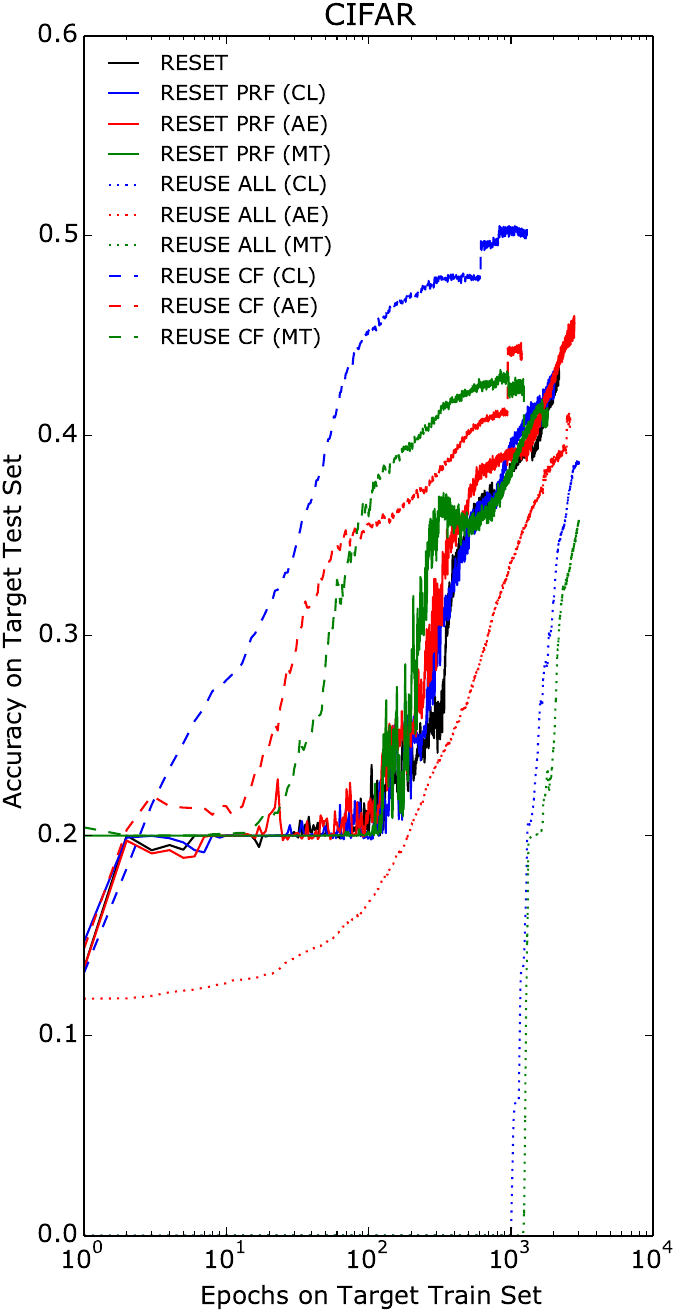}
\includegraphics[height=1.15\textheight]{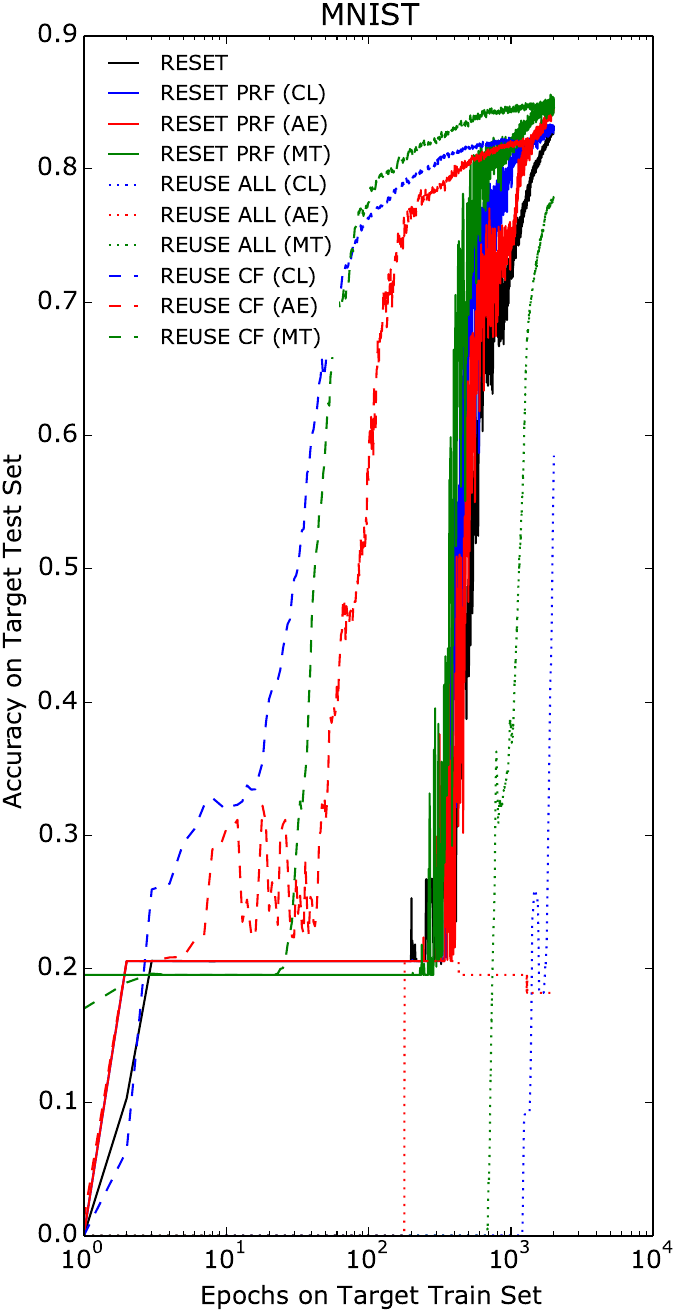}
\includegraphics[height=1.15\textheight]{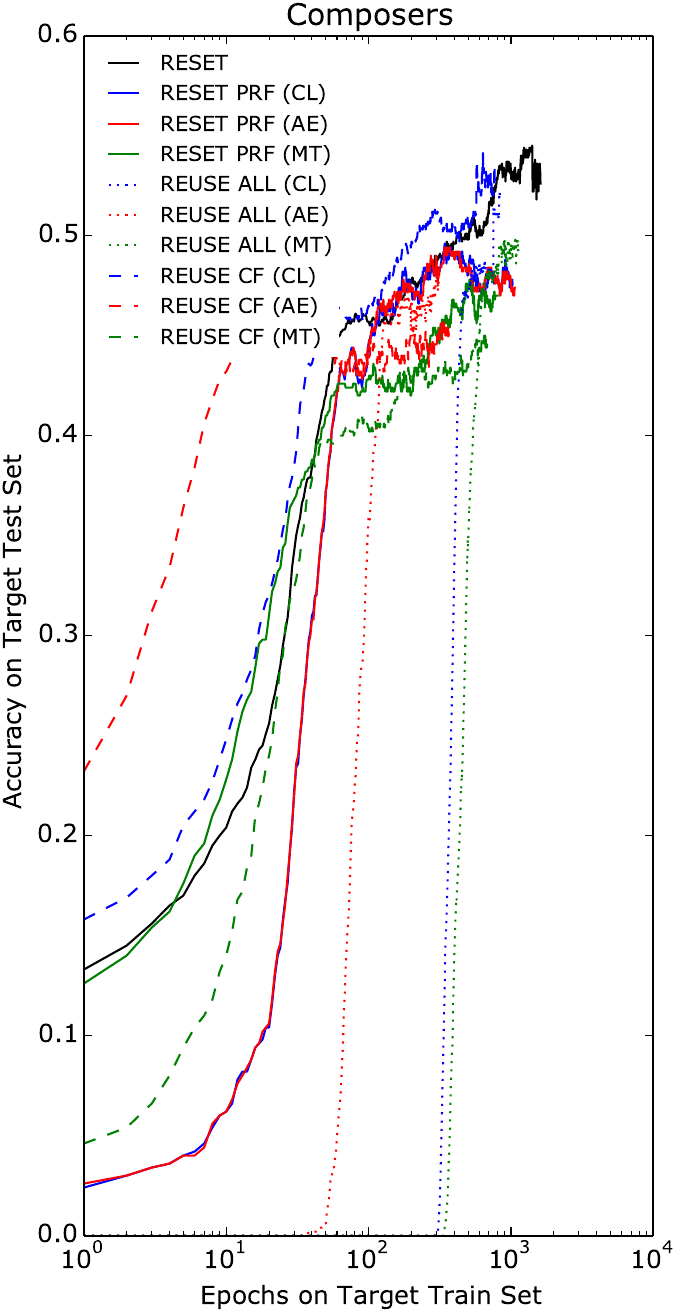}
\caption{Classification accuracy (proportion of correctly classified instances) on the target test-set of the different data sets, for different adaptation strategies.
Curves are averaged over two runs.
See table~\ref{tab:adapt-strategies} (page~\pageref{tab:adapt-strategies}) for an explanation of the legend}
\label{fig:acc}
\end{figure}
\end{landscape}

\begin{landscape}\centering
\begin{figure}
\includegraphics[height=1.15\textheight]{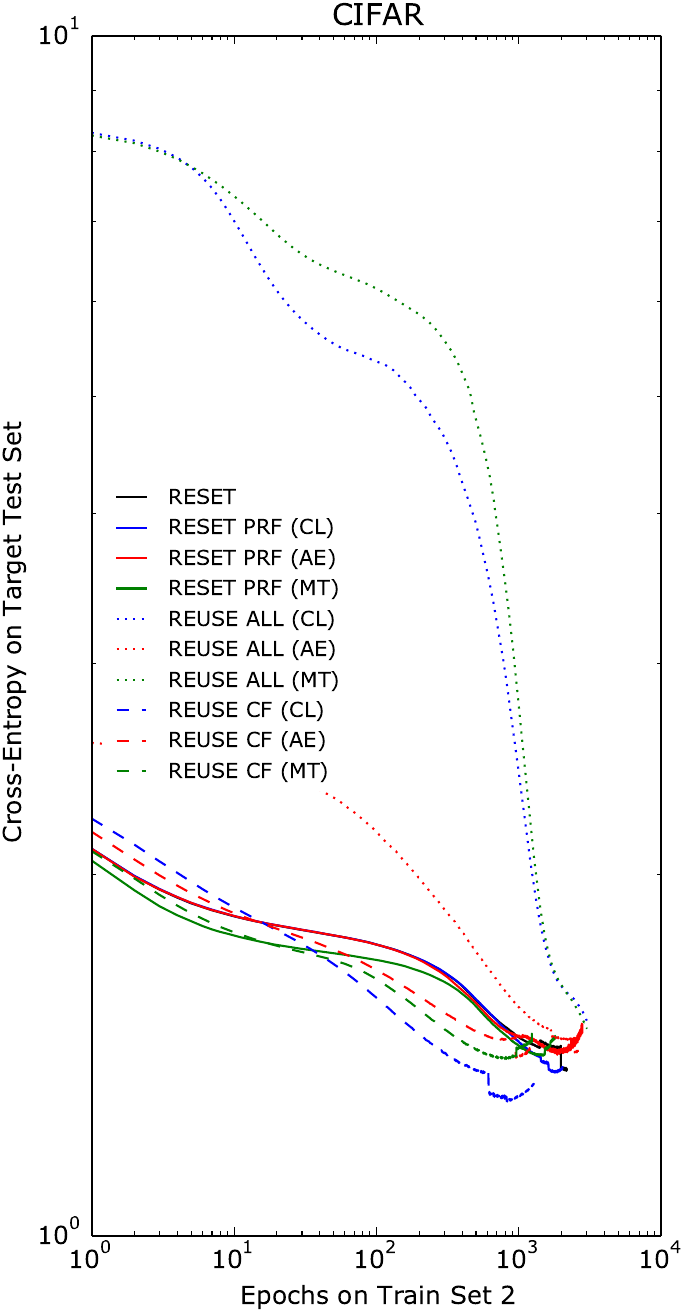}
\includegraphics[height=1.15\textheight]{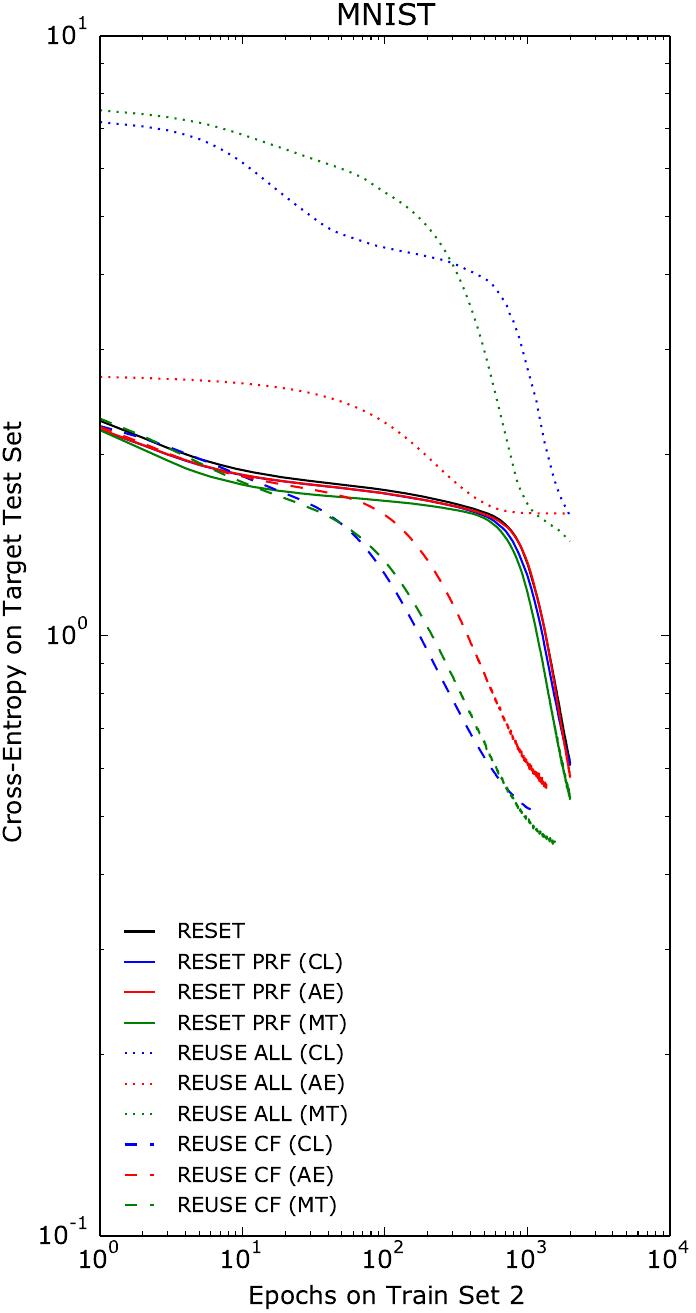}
\includegraphics[height=1.15\textheight]{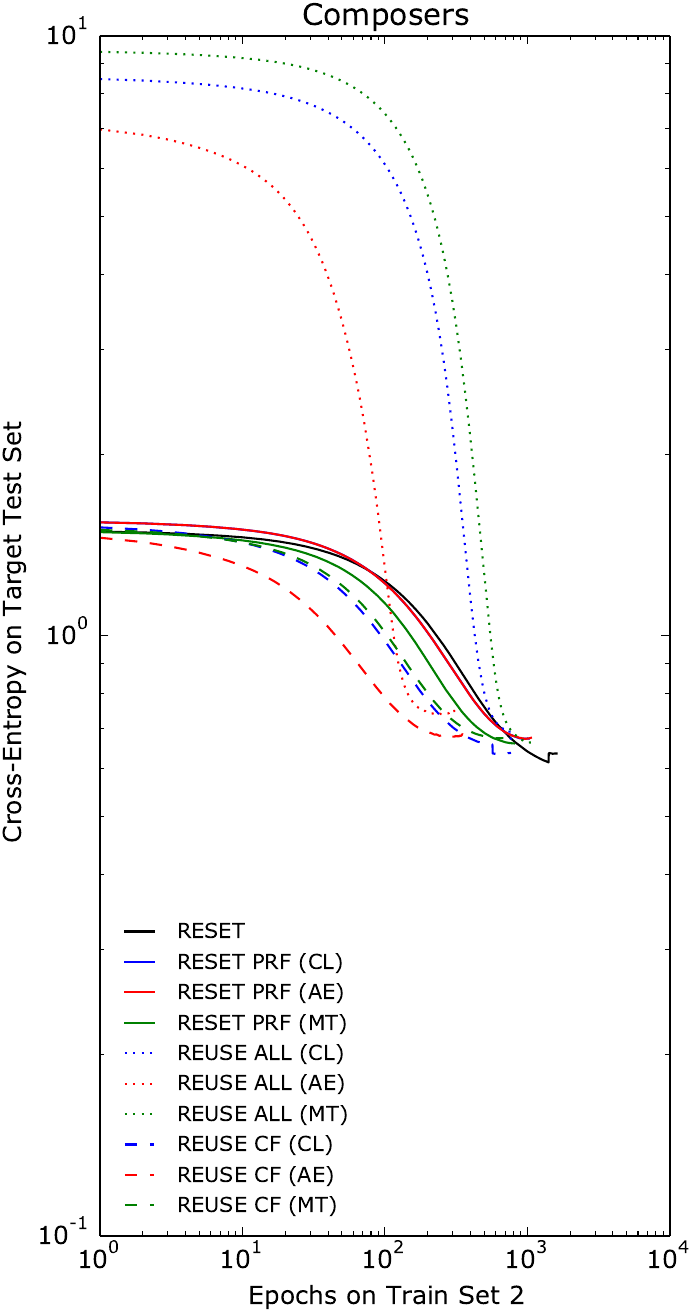}
\caption{Classification loss (categorical cross-entropy) on the target test-set of the different data sets, for different adaptation strategies.
Curves are averaged over two runs.
See table~\ref{tab:adapt-strategies} (page~\pageref{tab:adapt-strategies}) for an explanation of the legend}
\label{fig:clloss}
\end{figure}
\end{landscape}

\begin{landscape}\centering
\begin{figure}
\includegraphics[height=1.15\textheight]{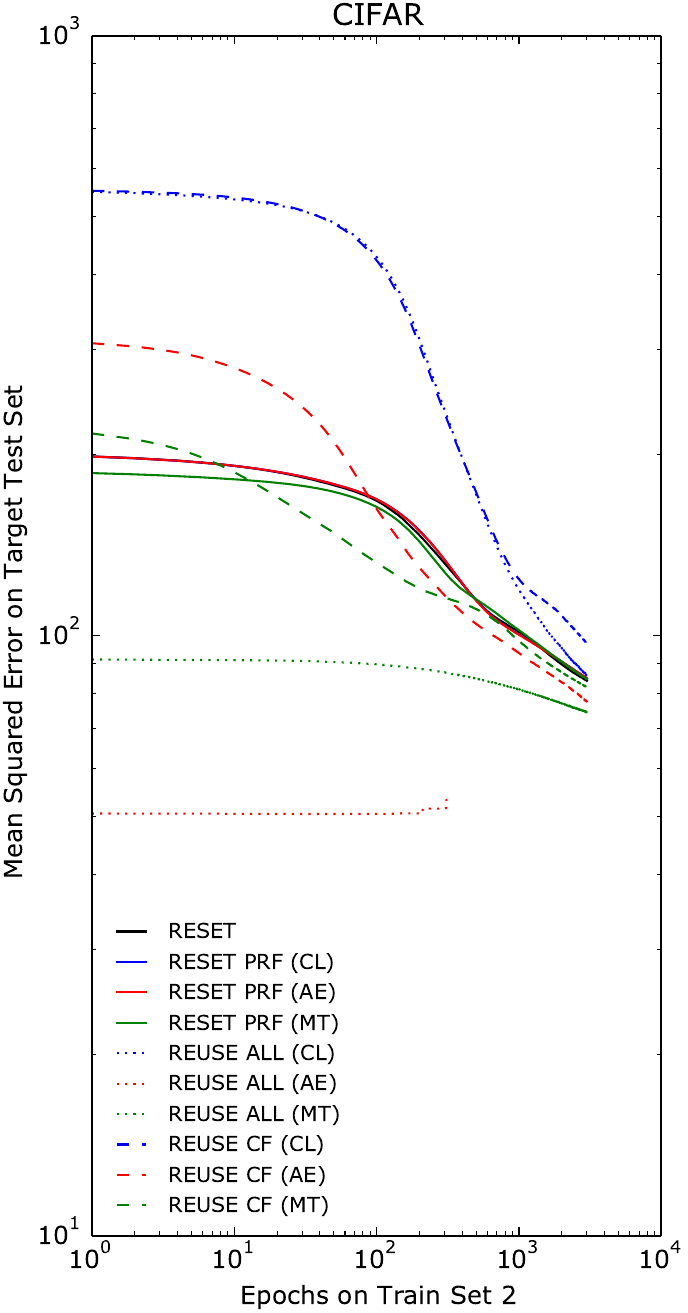}
\includegraphics[height=1.15\textheight]{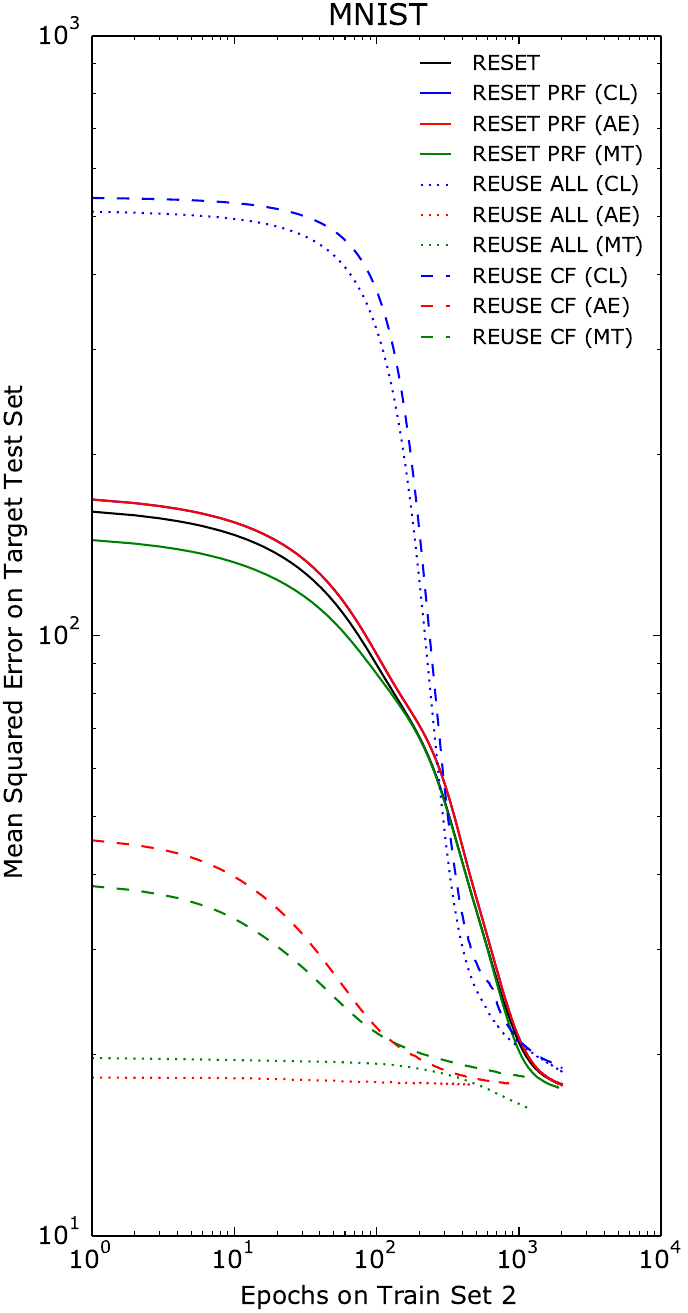}
\includegraphics[height=1.15\textheight]{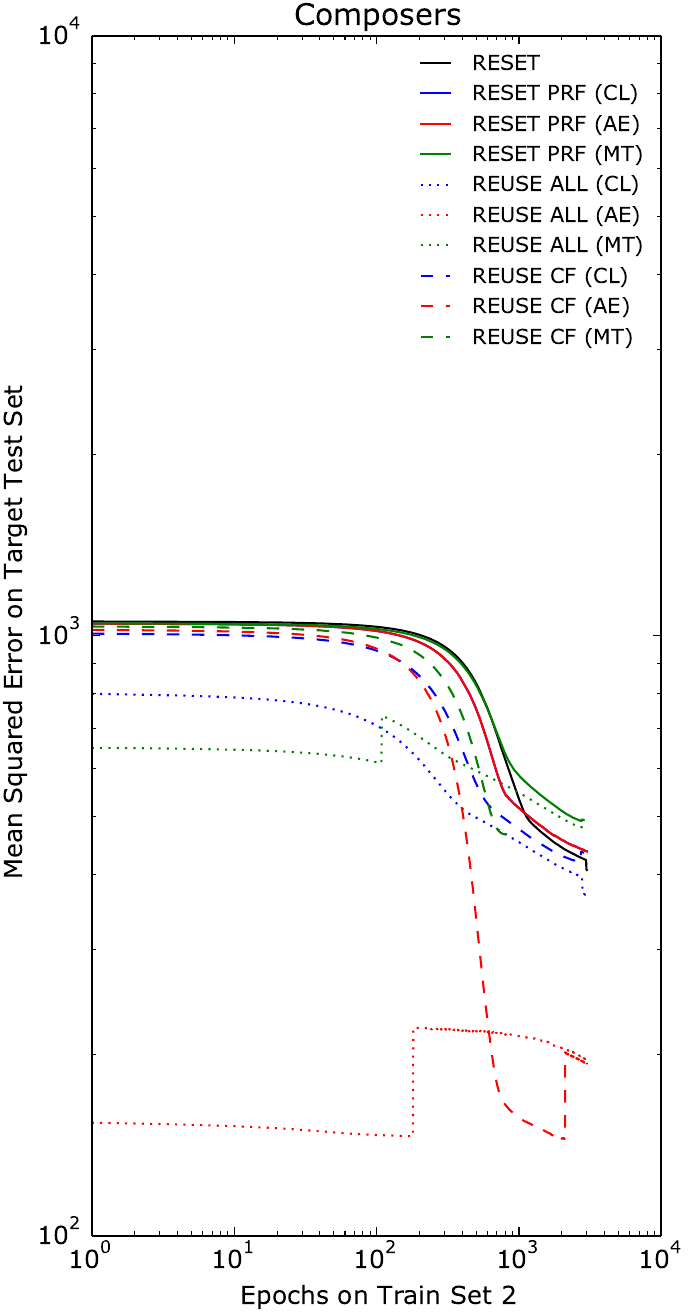}
\caption{Autoencoding loss (mean squared error) on the target test-set of the different data sets, for different adaptation strategies.
Curves are averaged over two runs.
See table~\ref{tab:adapt-strategies} (page~\pageref{tab:adapt-strategies}) for an explanation of the legend}
\label{fig:aeloss}
\end{figure}
\end{landscape}

\section{Conclusions}\label{sec:conclusions}
When interacting with a dynamical and unpredictable environment, the ability of an agent to easily adapt the conceptual space with which it interprets the environment, is a strong advantage.
As argued in Section~\ref{sec:from-adapt-repr}, this adaptability may also be one of the underlying mechanisms of creative behaviour.
In this document, we have described and evaluated several adaptation strategies to transform learned representations optimized for a specific task in a specific domain (the \emph{source} task/domain), to another task in another domain (the \emph{target} task/domain).
We have evaluated the strategies using a convolutional neural network, on multiple data sets, comprising natural images, handwritten digits, and classical music.

The results show that, across domains and tasks, adaptation strategies that transform existing representations allow for a quicker adaptation to a new task in a new domain than starting the representation learning from scratch.
Although there is no single adapatation strategy that is universally superior to others, some clear patterns do emerge from the results.
Firstly, when the target task is classification, then a successful adaptation strategy is to keep the first level convolutional filters (i.e.
the lower level representations) from the source task/domain, and reset the rest of the parameters.
This strategy is even beneficial when the source task is autoencoding, rather than classification.
Secondly, for the autencoding target task, the reuse of the full model (rather than just the convolutional filters) is a successful strategy, in the sense that even from the start, the task performance with that strategy is comparable with the asymptotic performance of other strategies.
But in contrast to the other strategies, this strategy does not substantially improve the model beyond the initial performance.
The experiments do not provide evidence for a strong advantage of the prior regularization of the convolutional filters (Section~\ref{sec:prior-regularisation}), over the condition where the model is trained from scratch with only standard regularization.

More elaborate experiments are required to provide firmer conclusions on the merits of each of the adaptation strategies.
In particular, a longer training phase is necessary to provide better insight in the asymptotic behaviour of the strategies.

\section*{Acknowledgement}\label{sec:acknowledgement}
The project Lrn2Cre8 acknowledges the financial support of the Future and Emerging Technologies (FET)
programme within the Seventh Framework Programme for Research of the European Commission, under FET
grant number 610859.

\bibliographystyle{apalike}
\bibliography{bib/bib_mg,bib/bib_cc,bib/bib_sl}

\newpage
\appendix
\section{Activation and Loss functions}\label{sec:act_functions}
In this section, we provide definitions of the nonlinear activation and loss functions mentioned in the main text.
\subsection{Nonlinear activation functions}
The follow nonlinear activation functions are defined as scalar functions as $f \colon \mathbb{R} \mapsto \mathbb{R}$.
Following conventions in the machine learning literature~\cite{Bishop:2006}, when applied to general tensors (i.e., vectors, matrices or higher order tensors), these functions take the form $f \colon \mathbb{R}^{N_1\times \dots \times N_p} \mapsto \mathbb{R}^{N_1\times \dots \times N_p}$, with the function being applied elementwise.
\begin{enumerate}
\item Sigmoid
\begin{equation}
\sigma(y) = \frac{1}{1 + \exp(-y)}
\label{eq:sigmoid}
\end{equation}
\item Hyperbolic tangent
\begin{equation}
\tanh(y) = \frac{\exp(y) - \exp(-y)}{\exp(y) + \exp(-y)} = 2 \sigma(2 y) - 1
\label{eq:tanh}
\end{equation}
\item Rectified Linear Units
\begin{equation}
\text{ReLU}(y) = \left\{ \begin{array}{lll}y & &\mbox{for $y>0$}\\0 & &\mbox{otherwise} \end{array} \right.
\label{eq:relu}
\end{equation}
\item Softmax. Given an input $\mathbf{y}\in \mathbb{R}^{N_1\times \dots \times N_p}$, the softmax activation function is given by
\begin{equation}
\text{Softmax}(y_{j_1\dots j_p}) = \frac{\exp(y_{j_1\dots j_p})}{\sum_{k_1=1}^{N_1}\dots \sum_{k_p}^{N_p}  \exp(y_{k_1\dots k_p})}\label{eq:softmax},
\end{equation}
where $y_{j_1\dots j_p}$ is the $j_1\dots j_p$-th element of $\mathbf{y}$.
\end{enumerate}
\subsection{Loss functions}\label{sec:cost_functions}
Let $\mathbf{X}=\{\mathbf{x}_1, \dots, \mathbf{x}_N\}$ be a set of inputs,  whose corresponding outputs are given by $\mathbf{y}(\mathbf{x}_i; \params)$ and $\mathbf{T}=\{ \mathbf{t}_1, \dots, \mathbf{t}_N\}$ a set of targets, with both $\faty, \mathbf{t} \in \mathbb{R}^{N_1\times \dots \times N_p}$.
\begin{enumerate}
\item Mean Squared Error
\begin{equation}
L_{MSE}(\params) = \sum_{n=1}^N\norm{\mathbf{t}- \mathbf{y}(\mathbf{x}_n ; \params)}_2^2
\label{eq:MSE}
\end{equation}
\item Categorical Cross Entropy
\begin{equation}
L_{CCE}(\params) = - \frac{1}{N}\sum_{n=1}^N \sum_{k_1=1}^{N_1} \dots \sum_{k_p=1}^{N_p} t_{n;k_1\dots k_p} \log\left(y_{k_1\dots k_p}(\mathbf{x}_n ; \params)\right),
\label{eq:CCE}
\end{equation}
where $t_{n;k_1\dots k_p}$ represents the $k_1\dots k_p$-th component of the $n$-th element of set $\mathbf{T}$.
%

\end{enumerate}
\end{document}